\documentclass[]{article} 
\usepackage{graphicx}
\usepackage{amssymb}
\usepackage{url,hyperref}
\usepackage[utf8]{inputenc}
\usepackage[T1]{fontenc}    

\usepackage{microtype}
\usepackage{subfigure}
\usepackage{booktabs} 
\usepackage{enumitem}
\usepackage{seqsplit, makecell, xltabular}

\usepackage{amsmath}
\usepackage{amssymb}
\usepackage{mathtools}
\usepackage{amsthm}
\usepackage{bbm}
\usepackage{bm}
\usepackage{multirow}
\usepackage{makecell}
\usepackage{caption}
\usepackage{xcolor}

\usepackage{geometry}
\title{\textbf{Multi-channel learning for \\ integrating structural hierarchies into \\ context-dependent molecular representation}}

\author{Yue Wan$^1$, Jialu Wu$^2$, Tingjun Hou$^2$\thanks{Corresponding authors. \newline \textit{E-mail addresses}: tingjunhou@zju.edu.cn (T. Hou), kimhsieh@zju.edu.cn (C.-Y. Hsieh), xiaowei@pitt.edu (X. Jia)}, Chang-Yu Hsieh$^2$\footnotemark[1], Xiaowei Jia$^1$\footnotemark[1] \\ \textit{University of Pittsburgh, Department of Computer Science, Pittsburgh, PA 15260, United States}$^1$ \\
\textit{Innovation Institute for Artificial Intelligence in Medicine of Zhejiang University,} \\ 
\textit{College of Pharmaceutical Sciences, Zhejiang University, Hangzhou, 310058, China}$^2$
}
\date{}

\begin{document}
\maketitle
\thispagestyle{empty}

\begin{abstract}
Reliable molecular property prediction is essential for various scientific endeavors and industrial applications, such as drug discovery. However, the data scarcity, combined with the highly non-linear causal relationships between physicochemical and biological properties and conventional molecular featurization schemes, complicates the development of robust molecular machine learning models. Self-supervised learning (SSL) has emerged as a popular solution, utilizing large-scale, unannotated molecular data to learn a foundational representation of chemical space that might be advantageous for downstream tasks. Yet, existing molecular SSL methods largely overlook chemical knowledge, including molecular structure similarity, scaffold composition, and the context-dependent aspects of molecular properties when operating over the chemical space. They also struggle to learn the subtle variations in structure-activity relationship. This paper introduces a multi-channel pre-training framework that learns robust and generalizable chemical knowledge. It leverages the structural hierarchy within the molecule, embeds them through distinct pre-training tasks across channels, and aggregates channel information in a task-specific manner during fine-tuning. Our approach demonstrates competitive performance across various molecular property benchmarks and offers strong advantages in particularly challenging yet ubiquitous scenarios like activity cliffs.
\end{abstract}

\section*{Introduction}
Empowered by the advancement of machine learning techniques, molecular machine learning has shown its great potential in computational chemistry and drug discovery \cite{goh2017deep, deepchem}. The data-driven protocol allows the model to infer biochemical behaviors from simple representations like SMILES sequence \cite{smiles} and molecular graph, enabling fast identification of drug candidates via rapid screening of vast chemical spaces \cite{virtual_screen}, as well as prediction of binding affinity, toxicity, and other pharmacological properties \cite{MoleculeNet, MoleculeACE}. These advancements significantly accelerate the drug discovery procedures, saving time and efforts from the traditional wet-lab experiments \cite{wetlab1, wetlab2}. However, it is fundamentally challenging to learn an effective and robust molecular representation via machine learning, limited by the expensive gathering of precise biochemical labels and the complexity underlying the structure-property relationships (SPR). With data scarcity, models may become overly adapted to specific structural patterns within the training molecules, making it fail to generalize to the broader chemical space. In addition, the challenge of ``activity cliffs'' \cite{AC} in drug discovery, where minor changes in molecular structure significantly alter the biological activity, further impose obstacles in developing accurate Quantitative SPR (QSPR) models \cite{MoleculeACE, ac_challenge1, ac_challenge2, Deng2023}. Activity cliffs refer to the concept where structurally similar molecules may exhibit significantly different biological activities. Understanding activity cliffs is crucial for drug discovery, as it enables more efficient lead optimization and enhances predictive modeling with better identification and development of potential drug candidates \cite{understand_ac, understand_ac2}.

Inspired by the success of the pretrain-finetune workflow in computer vision \cite{imagenet} and natural language processing \cite{bert, gpt2}, various methods in molecule self-supervised learning (SSL) \cite{gnnpretrain1, gnnpretrain_grover, gnnpretrain_molebert, molclr, graphlog, dsla, kano, gem} have emerged. 
In the self-supervised setting, machine learning models are pre-trained to learn generic molecular representations by optimizing the performance on pre-defined tasks on large-scale unannotated molecule data. These tasks are designed in a way such that solving them requires identification of important structural patterns and understanding of rudimentary chemical knowledge. Existing molecule SSL methods can be mainly classified into two categories: predictive and contrastive. Predictive learning \cite{gnnpretrain1, gnnpretrain_pred, gnnpretrain_grover, gnnpretrain_molebert, uni-mol, molformer} aims to predict structural components given contexts at different levels, which mainly focuses on intra-data relationship. These methods often follow the conventional pipeline of reconstructing the molecular information from masked inputs. Contrastive learning \cite{gnnpretrain_4_contrast, gnnpretrain_contrast_5, gnnpretrain_contrast_6, gnnpretrain_graphcl, gnnpretrain_molebert, graphlog, dsla}, initially proposed in computer vision \cite{simclr}, aims to learn the inter-data relationship by pulling semantic-similar data samples closer and pushing semantic-dissimilar samples apart in the representation space. Note that this idea aligns well with common heuristics in chemistry, where structurally similar molecules are likely to exhibit similar physicochemical and biological properties. Most works attempt to generalize the same SSL methods across multiple domains (e.g., social networks and molecular graphs), whereas it is unsure whether the same learning schemes are compatible to all settings. Recently, several studies have pointed out that existing SSL methods may fail to learn effective molecular representation. RePRA \cite{pretrainhelpornot} was proposed to measure the representations' potential in solving activity cliffs and scaffold hopping \cite{SH}. These are challenging yet ubiquitous tasks in drug discovery that requires the model to understand subtle chemical knowledge behind SPR. Experiments showed that most pre-trained representations perform worse than molecular fingerprints. The work in \cite{gnnpretrainhelpornot} explored various molecular SSL methods and observed that some pre-training strategies can only bring marginal improvement, while some may induce negative transfer. Meanwhile, studies have also highlighted the incompetence of several pre-trained representations in predicting binding potency under activity cliffs \cite{Deng2023, MoleculeACE, wu2023instructbio}. 

This work aims to enhance molecular representation learning that encodes robust and generalizable chemical knowledge. We start by identifying the two major drawbacks in existing methods: Firstly, in contrastive learning, the conventional formulations of the semantic-similar/dissimilar (i.e., positive/negative) samples are not well-tailored for molecular graphs. Most graph contrastive methods generate positive samples via graph perturbation, such as node/edge addition/deletion \cite{graphlog, dsla, molclr, gnnpretrain_molebert, molclr}. However, when applied on molecular graphs, chemical validity may be easily challenged. Molecules may also lose essential characteristics by perturbing important motifs (e.g., breaking an aromatic ring), shifting the ``semantics'' distant away. The negative samples (i.e., different molecules) are often treated equally, which essentially neglects the molecule structural relationship and the presence of specific molecular components; Second, almost all existing works attempt to learn a context-independent molecular representation space, aiming to generalize to various applications. However, this contradicts the fact that molecular properties are often context-dependent, from both the physical (e.g., surrounding environments) and biological (e.g., interaction with proteins) perspectives. In other words, it remains uncertain whether the same SSL tasks could align well with diverse downstream tasks of distinct properties in fine-tuning, thereby leading to the learning gap.

To approach the aforementioned challenges, we introduce a prompt-guided multi-channel learning framework for molecular representation learning. Each of the $k$ channels, guided by a specific prompt token, is responsible for learning one dedicated SSL task. Essentially, the pre-trained model is able to learn $k$ distinct representation spaces. During fine-tuning, a prompt selection module aggregates $k$ representations into a composite representation and uses it for the downstream molecular property predictions. This involves determining which information channel is most relevant to the current application, thereby making the representation context-dependent. We later show how this composite formulation is more resilient to label overfitting and manifests better robustness. In addition, we design the pre-train tasks to form an interpolation from a global view to a local view of the molecular structures. Besides leveraging the global molecule contrastive learning and the local context prediction \cite{gnnpretrain_grover}, we introduce the task of scaffold contrastive distancing, highlighting the fundamental role of scaffolds in affecting molecular characteristics and behaviors. Since scaffolds are often treated as starting points for new compound design, scaffold distancing aims to map molecules with similar scaffolds (generated via \textit{scaffold-invariant perturbations}) closer in the representation space. Additionally, it pushes molecules with different scaffolds apart, where the distance margin is computed adaptively based on structure composition difference. Note that scaffold distancing tackles the partial but core view of the molecules. The overall framework is pre-trained using ZINC15 \cite{zinc}, and evaluated on 7 molecular property prediction tasks in MoleculeNet \cite{MoleculeNet} and 30 binding potency prediction tasks in MoleculeACE \cite{MoleculeACE}. Learning to leverage information from different channels for different applications, our method surpasses various representation learning baselines in both benchmarks. More importantly, our method is shown to handle the challenge of activity cliffs more effectively, whereas competing approaches are more susceptible to the negative transfer, leading to a substantial performance decline. This suggests that these methods may rely more on surface-level patterns even after pre-training or are more susceptible to knowledge forgetting during fine-tuning, causing them to struggle with challenging problems that require a nuanced understanding of chemical knowledge. On the contrary, our learned representation demonstrates enhanced ability in preserving pre-trained knowledge during fine-tuning, offering improved transferability and robustness compared to other baselines. Case study shows that our method has the potential to identify crucial patterns that contribute to activity cliffs, even when relying solely on topological information.

\section*{Results}

The proposed framework (Figure~\ref{fig:framework}) comprises three major components that differ from the conventional pretrain-finetune paradigm on molecules: (1) The prompt-guided multi-channel learning, (2) contrastive learning with adaptive margin, and (3) scaffold-invariant molecule perturbation. It demonstrates effectiveness on both the molecular property prediction \cite{MoleculeNet} and binding potency prediction \cite{MoleculeACE} benchmarks, offering enhanced robustness and interpretability.

\subsection*{Prompt-guided multi-channel learning} 
\begin{figure}[!ht]
    \centering
    \includegraphics[width=1\columnwidth, trim={1.6cm 9.8cm 2.4cm 1cm},clip]{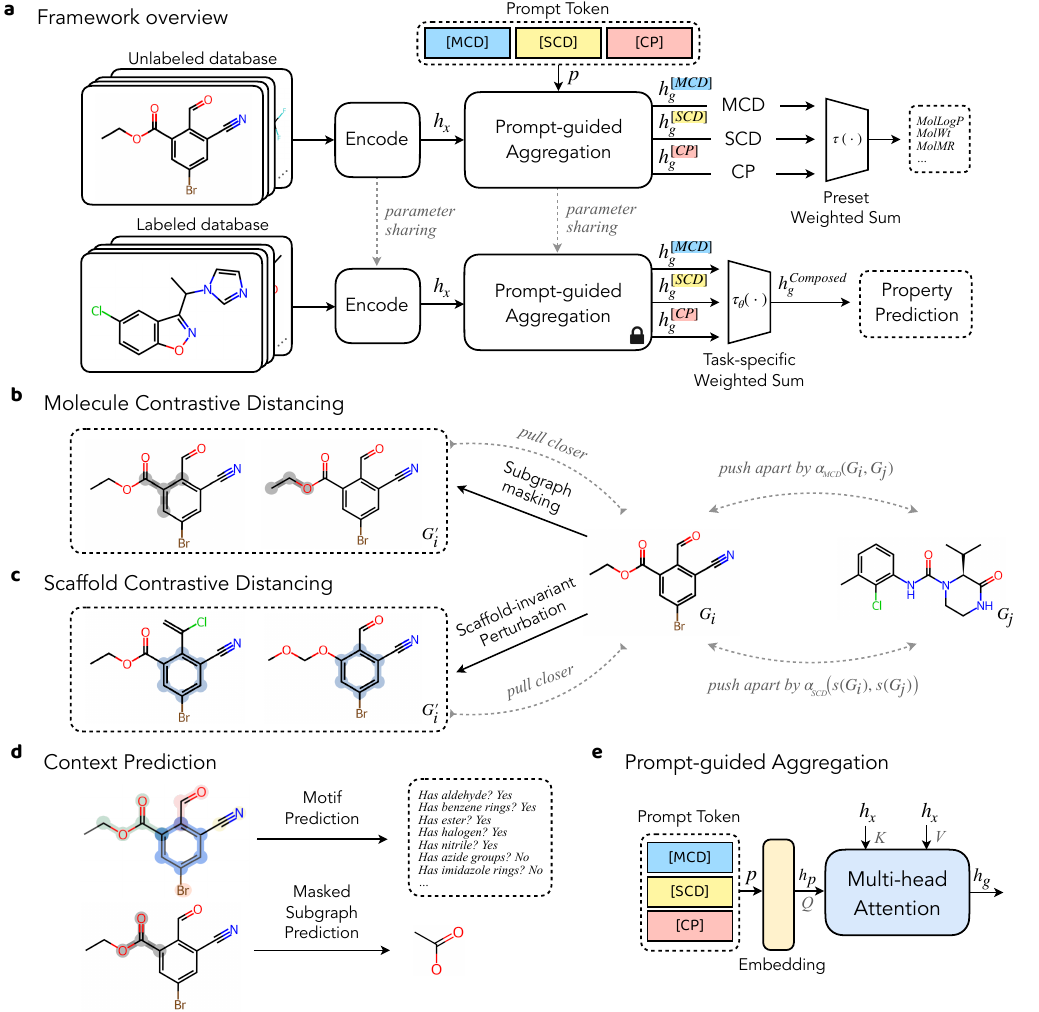}
    \caption{\textbf{Framework overview. a)} The prompt guided pretrain-finetune framework. For each downstream task, the model is optimized additionally on the prompt weight selection, locating the best pre-trained channel compatible with the current application. \textbf{b)} Molecule contrastive learning (MCD), where the positive samples $G_i'$ from subgraph masking is contrasted against negative samples $G_j$ by an adaptive margin. \textbf{c)} Scaffold contrastive distancing (SCD), where the positive samples $G_i'$ from scaffold-invariant perturbation is contrasted by against negative samples $G_j$ by an adaptive margin. \textbf{d)} Context prediction (CP) channel consists of masked subgraph prediction and motif prediction tasks. \textbf{e)} Prompt-guided aggregation module, which conditionally aggregates atom representations into molecule representation by prompt token. It is realized via a multi-head attention with prompt embedding $h_p$ being the query.}  
    \label{fig:framework}
\end{figure}

We introduce a prompt-guided multi-channel learning framework for molecular representation learning, as shown in Figure~\ref{fig:framework}a. Essentially, the molecular graphs will first go through a unified encoder module, and then diverge into $k$ different channels, each of which is responsible for learning distinct SSL tasks. For each channel, a prompt token $p_i$ is utilized to distinguish levels of molecule representation. This is realized via a prompt-guided readout operation \cite{graphprompt}, which aggregates atom representations conditionally into molecule representation given the prompt token (Figure~\ref{fig:framework}e). Our experiment involves three learning channels, which are molecule distancing, scaffold distancing, and context prediction. Each channel focuses on a unique aspect of the molecular structure, enabling molecular representation learning from a set of hierarchical viewpoints, from a global view (i.e., entire molecule), a partial view (i.e., core structure), and down to a local view (i.e., functional groups). 

We now briefly introduce the three learning channels. (i) Molecule distancing (Figure~\ref{fig:framework}b) is archived using a variant of triplet contrastive loss \cite{triplet}. It considers triplet of molecule samples \{anchor, positive, negative\}, where negative (i.e., dissimilar) samples are pushed apart against the anchor and positive (i.e., similar) samples by a margin $\alpha$. On top of this, we propose the \textit{adaptive margin} $\alpha{(\cdot)}$, as detailed below. It introduces another level of distancing constraints based on the structural similarity of molecule composition. We follow the work of Molecular Contrastive Learning of Representations (MolCLR) \cite{molclr} and generate positive samples via molecule subgraph masking. (ii) Scaffold distancing (Figure~\ref{fig:framework}c) is a contrastive learning task that focuses on scaffold differences. Molecule scaffolds are viewed as the foundation for a range of biologically active molecules. They play a crucial role in drug discovery and medicinal chemistry by providing a starting point for compound designs with desired pharmacological properties. In other words, molecules with similar scaffolds are more likely to possess similar physical (e.g., solubility, lipophilicity) and biological (e.g., conformational property when interacting with a protein) characteristics, and thereby sharing similar semantics. Scaffold distancing contrasts the \textit{scaffold-invariant molecule perturbations}, as detailed below, against molecules with different scaffolds using the \textit{adaptive margin loss}. (iii) Context prediction (Figure~\ref{fig:framework}d) involves masked subgraph prediction and motif prediction, which are also adopted in GROVER \cite{gnnpretrain_grover}. For each molecular graph, a random subgraph (i.e., a center atom and its one-hop neighbors) is masked out, and the model aims to reconstruct the subgraph based on its surrounding structures. Motif prediction aims to predict the existence of functional groups within the molecule. This learning channel mainly focuses on the local view of the molecule by identifying the existence of substructure and functional groups, while molecule distancing and scaffold distancing focus more on the global view and partial view, respectively. Each channel has its own readout layer, termed prompt-guided node aggregation (Figure~\ref{fig:framework}e). Presumably, the distribution of atom importance should be different across channels (Supplementary Figs. 7-9 and Supplementary Note 6). The learned channel-wise representations exhibit a high correspondence to the associated structural features (Supplementary Fig. 1 and Supplementary Note 1) and support hierarchical iterative clustering of the chemical space (Supplementary Fig. 2 and Supplementary Note 2). To further improve the robustness of the learned knowledge under this framework, we incorporate two regularization tricks on the intra-channel node aggregation and the inter-channel alignment. For the molecule and scaffold channels, the aggregation attentions are more encouraged to span across all atoms and scaffold atoms, respectively. Meanwhile, a small set of supervised tasks is utilized to regularize the composite representation under three prompt weight presets, hence improving the channel alignment. A comprehensive ablation study is provided in Supplementary Fig. 10 and Supplementary Note 7.1. 

In the fine-tuning stage, the model is initialized with the same molecule encoder and prompt-guided aggregation modules, along with the learned parameters from pre-training. The parameters in the aggregation modules are fixed during fine-tuning as we aim to use it as a pooling layer independent of the downstream applications. In addition, a prompt-tuning module $\tau_{\theta}(\cdot)$ is introduced to determine which channel is most relevant to the current application. It essentially learns a task-specific prompt distribution. The prompt weights are utilized to linearly combine the channel-wise information into a composite molecule representation, which is then used for the task-specific prediction. We discover that this approach is more effective than simply concatenating the representations (Supplementary Fig. 11 and Supplementary Note 7.2). We initialize the prompt weights by choosing the candidate which leads to the smoothest quantitative structure-property landscape (i.e., lowest roughness index) \cite{roughness} of the composite representation. More details are provided in the Method section.

\subsection*{Contrastive learning with adaptive margin} We further introduce the adaptive margin loss, a variant of the triplet loss \cite{triplet}, that supports the contrastive learning in the first two channels (molecule distancing and scaffold distancing). In the conventional triplet loss, the representation distance between the anchor $G_i$ and negative (i.e., semantic-dissimilar) sample $G_j$ needs to be at least by margin $\alpha$ larger than the distance between the anchor $G_i$ and positive (i.e., semantic similar) sample $G_i'$. Note that this margin remains the same for any triplet considered. However, when applied to molecule triplets, it neglects the known structural relationship between molecules (e.g., co-existence of functional groups). To learn a more fine-grained molecule representation space, we propose to adaptively compute the molecule triplet margin based on the Tanimoto similarity between molecule fingerprints. As shown in Figure~\ref{fig:framework}b and c, the adaptive margin $\alpha_{\text{MCD}}(.)$ considers the molecule structural similarity between $G_i$ and $G_j$, while $\alpha_{\text{SCD}}(.)$ considers the scaffold structural similarity between $s(G_i)$ and $s(G_j)$. Another issue with the conventional triplet loss is that it imposes no constraint on the representation space beyond the margin. It means that the actual representation distances are not necessarily to be well correlated with the computed margin, even if the margin constraints are fully satisfied. This is further elaborated the example in Supplementary Fig. 3. Therefore, we include a secondary term into the adaptive margin loss by considering the structural relationship among the anchor and different negative samples. Detailed formulation of the adaptive margin loss is included in Method section. With careful consideration of existing structural similarity, the learned representation space would better capture molecule relationships in a fine-grained representation space. A performance drop is observed when the adaptive margin loss is replaced with the conventional margin loss (Supplementary Fig. 10 and Supplementary Note 7.1).

\subsection*{Scaffold-invariant molecule perturbation} 

To generate semantic-similar samples (i.e., positive) for scaffold contrastive distancing, we propose to perturb only the terminal side chains of the molecule. In other words, the molecule scaffold (i.e., core structure) is preserved. This is done by first identifying the side chains and then performing fragment replacement based on a candidate fragment pool. To avoid significant alterations in molecule characteristics, we restrict the amount of changes to be fewer than five atoms. For simplicity reason, we consider the Bemis-Murcko framework as the scaffold. Figure~\ref{fig:framework}c shows a sample perturbation with scaffolds highlighted in blue. In this example, the benzene ring is the identified scaffold, and either the carboxylic ester group or the carbonyl group is perturbed by another functional groups. Note that such perturbation is not limited to atom-level or bond-level editing, but also motif-level. 

\subsection*{Molecular property prediction} 

To demonstrate the effectiveness of our approach, we first evaluate it on seven challenging classification datasets from MoleculeNet \cite{MoleculeNet}, which is a large-scale curated benchmark that covers multiple molecular property domains (e.g., physiology, biophysics). The scaffold split scheme \cite{scaffoldsplit} is applied. The performance is evaluated using the ROC-AUC value. Each experimental result is averaged over three different runs following the prior works \cite{molclr, gnnpretrain_grover, gem}. To demonstrate the effectiveness of our framework across different model architectures, we pre-train both a graph neural network GIN \cite{gin} and a graph transformer GPS \cite{gps}, termed Ours$_{\text{GIN}}$ and Ours$_{\text{GPS}}$, respectively. We compare our methods with {twelve} competitive molecular representation learning baselines, which cover a wide variety of pre-train SSL techniques, model architectures, and input representations (i.e., sequence, graph, geometry).  Specifically, we first consider four contrastive baselines, including GraphLoG \cite{graphlog}, D-SLA \cite{dsla}, MolCLR \cite{molclr}, and KANO \cite{kano}. These methods also share the commonality of using GNNs to encode molecular graphs. Besides, we include six predictive baselines, including Hu et. al \cite{gnnpretrain1}, GROVER \cite{gnnpretrain_grover}, MoLFormer \cite{molformer}, KPGT \cite{kpgt}, GEM \cite{gem}, and Uni-Mol \cite{uni-mol}. MoLFormer serves as a strong baseline for sequence-based models, while GEM and Uni-Mol are strong baselines for 3D geometry models. We further incorporate GraphMVP \cite{graphmvp} and ImageMol \cite{imagemol} as two multi-task learning baselines whose pre-training tasks cover both contrastive and predictive learning. Last but not least, we train the GIN and GPS models from scratch. Table~\ref{tab:moleculenet} shows the performance comparison results. Our methods improve over the no-pretrain setting (i.e., GIN and GPS) by 12.6\% and 8\% ROC-AUC in average, respectively. When compared to the other SSL methods, our approach reaches the new state-of-the-art performance on BBBP, Clintox, BACE and SIDER datasets, while remaining highly competitive in the rest of the tasks. Our overall ROC-AUC score is 0.8\% higher than the second best method (Uni-Mol). Notably, Uni-Mol leverages 3D geometric information and is an order of magnitude larger (in term of model parameters) than both Ours$_\text{GIN}$ and Ours$_\text{GPS}$.

\subsection*{Binding potency prediction} 

We consider MoleculeACE \cite{MoleculeACE} as the second evaluation benchmark. It consists of 30 datasets retrieved from ChEMBL \cite{chembl}. Each dataset contains binding potency measures (e.g., $K_i$ value) of molecules against a macromolecular target. These datasets mainly focus on the structure-property relationships (SPR), where the phenomenon of activity cliffs is amplified. Activity cliffs refer to the cases where small changes in molecular structure significantly alters its biological activity, and understanding them is crucial for optimizing lead compounds and designing new molecules with desired activities. The phenomenon is also counter-intuitive in machine learning, as the machine learning model tends to make similar predictions given similar inputs. We take R-squared as the evaluation metric since relative binding potency ranking is more important than absolute prediction errors (e.g., RMSE) in the real world. 

\begin{figure}[!ht]
    \centering
    \includegraphics[width=1\columnwidth, trim={0.2cm 12.5cm 0cm 0cm},clip]{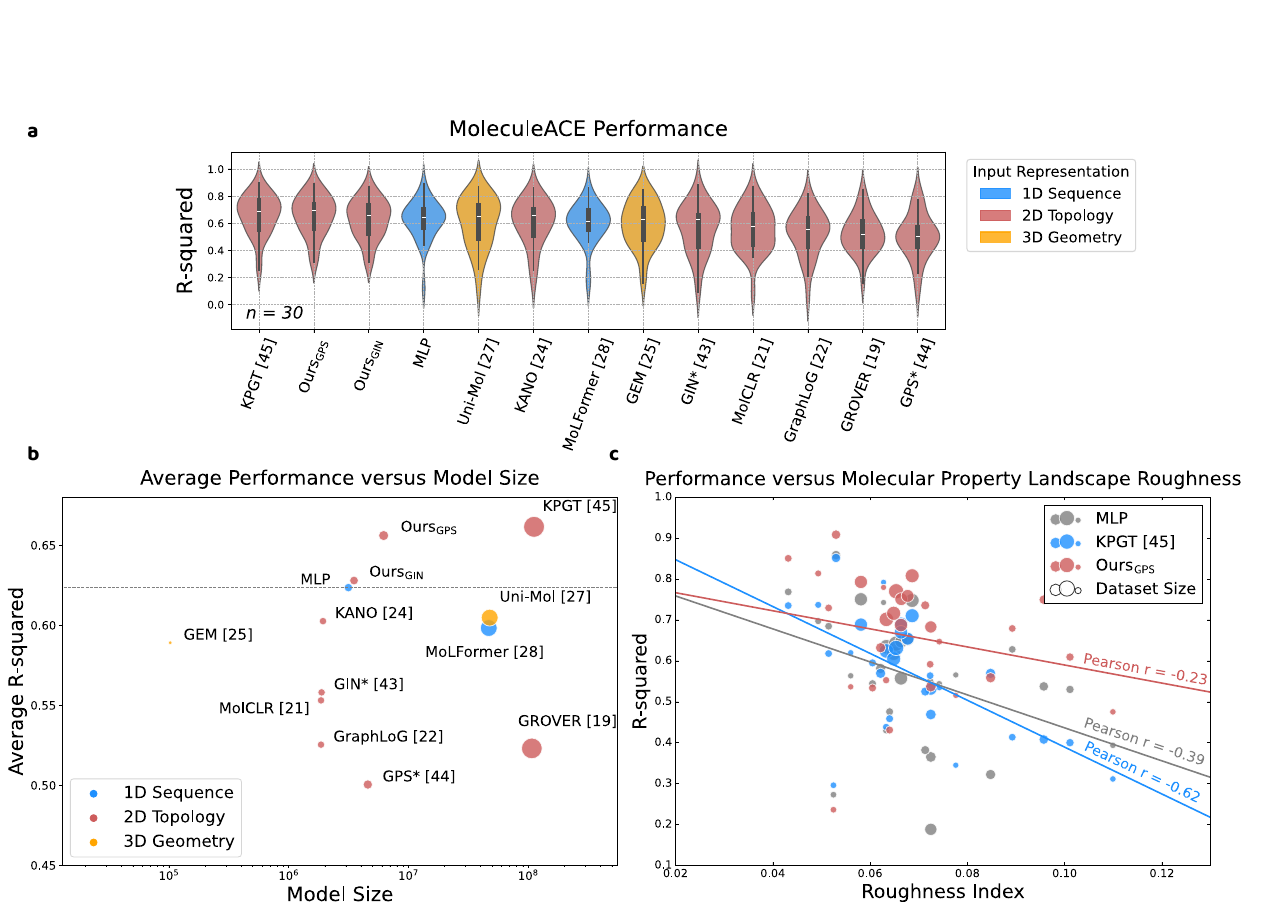}
    \caption{\textbf{Binding potency prediction. a)} The violin plot illustrates model performance across 30 binding potency prediction tasks in MoleculeACE \cite{MoleculeACE}. Performance on each dataset is averaged across three independent runs using stratified splits. The asterisk (*) indicates that the model is trained from scratch (no pre-train). The ordering of the methods from left to right is sorted by the average performance. The width of each violin represents the density of data points, with wider sections indicating a higher concentration of values. The minima and maxima are displayed by the lower and upper ends of the violin. The middle solid white line represents the median value. \textbf{b)} Average model performance across 30 binding potency prediction tasks with respect to model sizes (also indicated by the size of the dots). The dashed line follows the performance of the multi-layer perceptron (MLP), which serves as the baseline comparison between traditional fingerprints and deep molecular representations. \textbf{c)} {Relationship between model performance and the presence of activity cliffs in each dataset measured by the roughness of molecular property landscape. Performance on each dataset is averaged across three independent runs using random splits. The size of the datasets is indicated by the size of the dots.} The slope of the fitted line of each method indicates the correlation. Ours$_\text{GPS}$ is used in this analysis.}
    \label{fig:moleculeace}
\end{figure}

Figure~\ref{fig:moleculeace}a shows the performance of thirteen methods on the MoleculeACE benchmark under stratified splits \cite{MoleculeACE} (Supplementary Table 2). The multi-layer perceptron (MLP) is trained using the ECFP4 fingerprint. We also compare the average performance with respect to the model sizes (i.e., number of parameters) in Figure~\ref{fig:moleculeace}b, as indicated by its x-axis and the size of the dots, and types of input representations (blue = 1D sequence, red = 2D topological graph, and yellow = 3D geometry). In general, Ours$_\text{GPS}$ ranks in the second place, but with a significantly smaller model size than KPGT. Besides, almost all methods fail to surpass MLP with fingerprints. MolCLR and GraphLoG also show negative transfer compared to the GIN model. It demonstrates the incompetence of existing methods in learning the nuances of chemical knowledge behind the structural-property relationship, regardless of the input representations, pre-training strategies, and the model sizes \cite{MoleculeACE, Deng2023}. One of the reasons behind KPGT's strong performance could be attributed to the fact that the molecule fingerprint and descriptors are heavily embedded within the model. In addition, KPGT is pre-trained using the ChEMBL database \cite{chembl}, such that over 99\% of the testing molecules in MoleculeACE are already exposed to the model during the pre-training stage. In contrast, our models only have seen less than 5\% of the testing molecules during the pre-training stage. The plot also suggests that there are no clear advantages {of molecular representation learning using either 1D sequence or 3D geometry}, especially when comparing the performance with Uni-Mol and MoLFormer. 

{We further study the model performance in relation to the presence of activity cliffs in each dataset, which is measured by the roughness index (ROGI) \cite{roughness} with respect to the ECFP4 fingerprints and potency labels. A smaller ROGI value indicates a smoother structure-property landscape, hence less activity cliffs. Since ROGI is a quantitative SPR (QSPR) metric that analyzes the overall chemical space, it does not account for any distribution shift between chemical spaces. Hence, its correlation with model performance would be easily obscured by any distribution shift among training, validation, and testing sets (Supplementary Fig. 14). {For this reason, all experiments in this work that involves QSPR analysis are performed using random splits}, following the work in \cite{roughness}. As shown in Figure~\ref{fig:moleculeace}c and Supplementary Table 3, the model performance is negatively correlated with landscape roughness for all methods, which is consistent with the results in \cite{roughness}. Compared to the MLP with fingerprint and KPGT, our method is shown to be more robust against tasks with rough structure-property landscapes. In addition, the plot also indicates that our method performs better under data scarcity, where the size of dots represents the size of datasets.}

\subsection*{Representation robustness}
\begin{figure}[!ht]
    \centering
    \includegraphics[width=1\columnwidth, trim={0cm 8.8cm 0cm 0cm},clip]{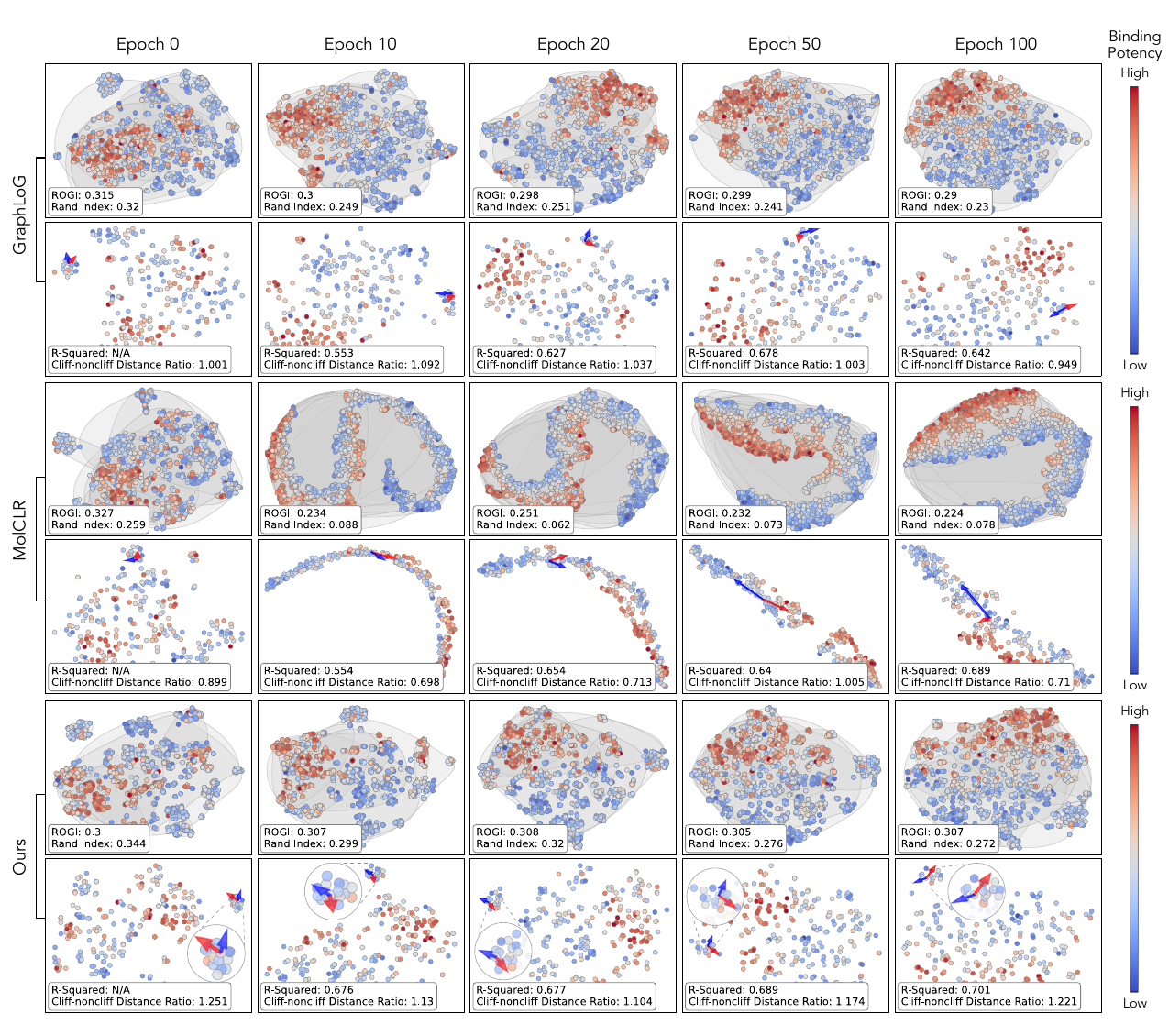}
    \caption{\textbf{Representation space probing.} The dynamics of molecule representation space of three methods at five finetune timestamps on dataset CHEMBL237\_Ki ($n=2602$). For each row pair, a 2D view of the representation space of both the training (top) and validation set (bottom) are visualized. The color map represents the normalized binding potency of each data point from lowest to highest within the dataset. Rand index and roughness index (ROGI) are reported for training, while R-squared value and cliff-noncliff distance ratio are reported for validation. The grey circles correspond to the clustering assignment using ECFP4 fingerprint. The red arrow indicates the distance of a cliff molecule pair, while the blue arrow indicates that of a non-cliff molecule pair.}
    \label{fig:probing}
\end{figure}

To study the robustness of the learned molecular representation, we propose to probe the fine-tuning process and evaluate the shift in the representation space. Essentially, the shift captures how much pre-trained chemical knowledge is distorted during fine-tuning. We examine the representation space of both the training and validation molecule set at five training timestamps. To clarify, the term ``learned representation'' refers to the numerical embedding used for the final prediction layer. We choose CHEMBL237\_Ki, one of the largest binding potency prediction dataset in MoleculeACE, for the downstream target. For fair comparison, this analysis is performed among Ours$_\text{GIN}$, GraphLoG, and MolCLR, as these methods only differ by the SSL strategies for pre-training, while using the same pre-train dataset, GNN architecture and hyperparameters. In Figure~\ref{fig:probing}, each column represents a training timestamp, and each row pair represents the visualization of representation space in training and validation sets. The coloring represents the normalized potency labels for illustrative purposes. We also report four additional metrics that capture the representation characteristics along the training process: 1. \textit{Roughness Index} (i.e., ROGI) \cite{roughness} captures the landscape roughness of molecular property given a representation. 2. \textit{Rand Index} \cite{rand_index} measures the proximity between fingerprint (ECFP4) clustering and representation clustering, serving as a proxy for the amount of structural information encoded within the current representation space. A higher Rand Index indicates a greater amount of structural knowledge being captured by the representation. 3. \textit{Cliff-noncliff Distance Ratio} indicates the generalizability of the representation towards the activity cliffs. In short, cliff matched molecule pairs (MMPs) (i.e., similar molecules with different labels) should be more distant away compared to the non-cliff MMPs (i.e., similar molecules with similar labels). It is calculated as the ratio of the average distance between cliff MMPs to that of non-cliff MMPs. We also show three molecules on the plot for illustrative purposes, with one cliff MMP (indicated by the red arrow) and one non-cliff MMP (indicated by the blue arrow). 4. At last, we report the validation \textit{R-squared} as the performance measure. Here are the three main takeaways from Figure~\ref{fig:probing}: 
\begin{itemize}
    \item Our composed representation yields the lowest ROGI value to start with. In other words, our pre-trained knowledge is more transferable to the target application. This is also shown by our rapid convergence rate in the validation set, reaching a validation R-squared of 0.676 at epoch 10 (Supplementary Fig. 5 and Supplementary Note 4).
    \item Our representation preserves the chemical knowledge learned from pre-training better than others, leading to less overfitting and better robustness. Since representations are optimized towards the property labels, the Rand index drops continuously for all methods. It means that the encoded information gradually shifts from being structure-oriented to label-oriented. However, our method has the lowest drop in Rand index of 0.072, compared to the drop of 0.09 by GraphLoG and 0.181 by MolCLR. The visualization also shows that the representations of MolCLR begin to overfit to the labels starting from epoch 10, resulting in the loss of substantial structural relationships between molecules. This explains its low ROGI value and validation R-squared along the training process.
    \item Our representation seems to exhibit a better understanding of the nuances of chemical knowledge in activity cliffs. Our average cliff-noncliff distance ratio in the validation set is always above one and larger than that of GraphLoG and MolCLR. As illustrated by the triplet samples, the red arrow is longer than the blue arrow across fine-tuning epochs, while the closeness of cliff and non-cliff pairs in space (i.e., structurally similar) is maintained. For MolCLR, even though the red arrow can be much longer than the blue one, the molecules are distant away. It means that MolCLR fails to capture the structural similarity aspect of the activity cliffs. {A detailed examination of the distance histogram is visualized in Supplementary Fig. 13.}
\end{itemize}

\begin{figure}[!ht]
    \centering
    \includegraphics[width=1\columnwidth, trim={0cm 16.5cm 0cm 2cm},clip]{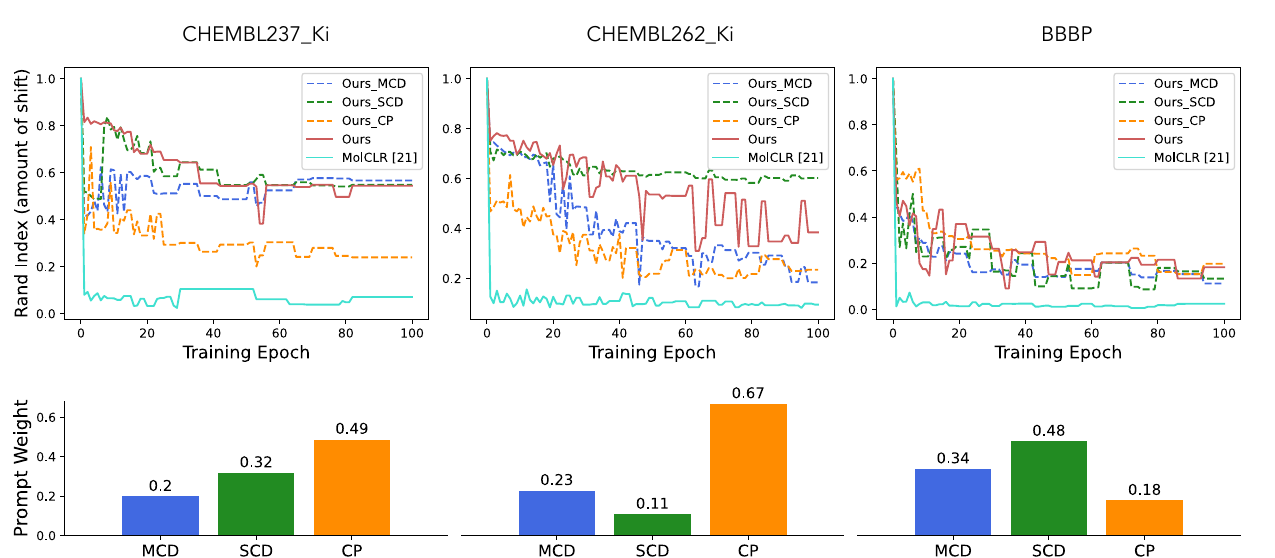}
    \caption{\textbf{Representation shift decomposition.} Detailed analysis of the representation shift during fine-tuning across three datasets. The dashed lines (top row) correspond to the representation shift (approximated by the Rand index) for each individual channel, while the solid lines represent the overall method. The bottom row displays the optimized prompt weights  distributed over channels.}
    \label{fig:rand_index}
\end{figure}

To further understand why our composed representation is more resilient to the representation shift and can better preserve pre-trained knowledge, we present more analysis on three diverse datasets, along with the individual channel-wise representation behavior during fine-tuning. We choose CHEMBL237\_Ki and CHEMBL262\_Ki as the representative regression-based datasets of different scales, and BBBP as the typical classification-based dataset. The Rand index of representation clustering difference between the initial and the current timestamp is computed as a proxy (slightly different than before) for the representation shift. A smaller Rand index indicates a larger shift. As illustrated in Figure~\ref{fig:rand_index}, channels with the highest prompt weights often exhibit the largest shift, and vice versa. This is reasonable because these channels contribute more to the optimization against the labels. Conversely, even though low-weighted channels contribute less to the optimization, they are more likely to preserve the pre-trained knowledge. As a result, the composed representation derived from channel aggregation exhibits a certain level of resilience to representation shifts, making it potentially more robust than other methods. The probing analysis under the few-shot learning settings also reveals similar patterns (Supplementary Fig. 6 and Supplementary Note 5).

\subsection*{Activity cliffs analysis}
We present a deeper analysis into the potential of our method in understanding the activity cliffs. To be more specific, we evaluate the relationship between the model explanations generated by the GNNExplainer \cite{gnnexplainer} and the predicted binding mode between the ligands and the protein pockets by AutoDock Vina \cite{trott2010autodock}. Note that this analysis merely serves as a proof of concept, such that our representation has the potential of capturing influential and well-established factors in binding affinities. However, the fundamental limitation of utilizing topological information only is unavoidable, which we will discuss in Conclusion.  

\begin{figure}[!ht]
    \centering
    \includegraphics[width=1\columnwidth, trim={1.5cm 13.6cm 1cm 1.3cm},clip]{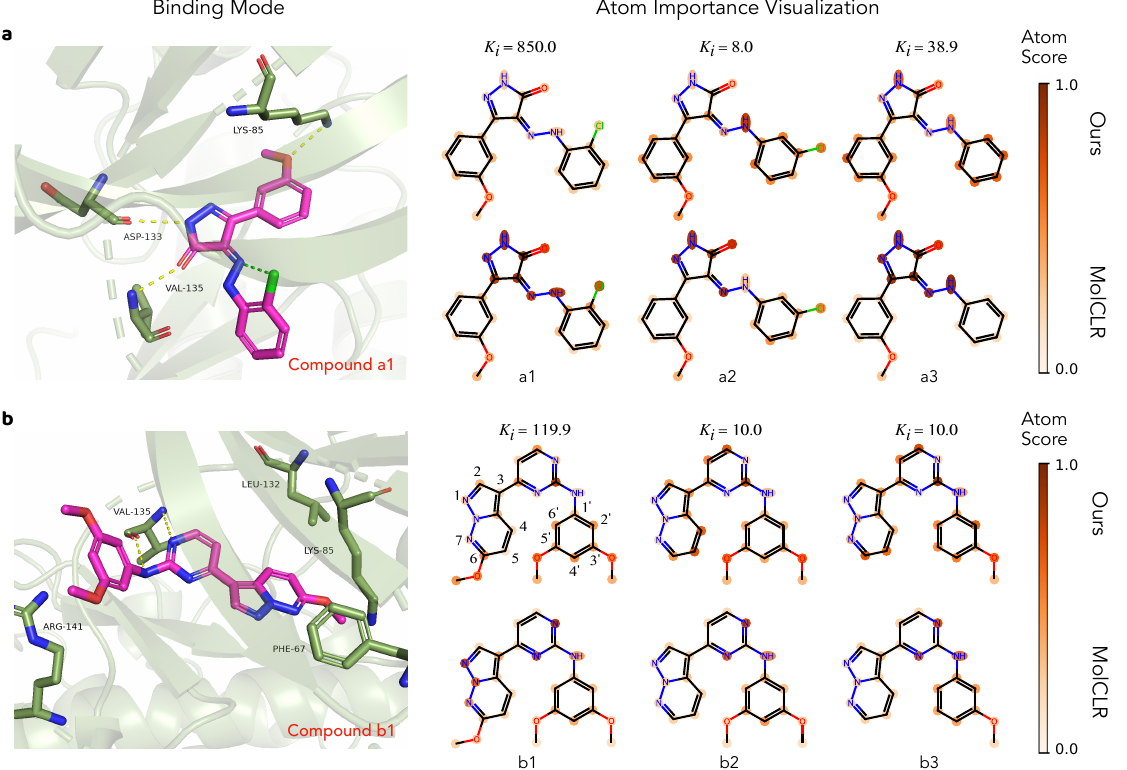}
    \caption{\textbf{Activity cliffs analysis.} Evaluate the bioactivity binding mode and the atom importance captured by our model on a (a) 5-phenyl-4-phenyldiazenyl-1,2-dihydropyrazol-3-one series and a (b) N-phenyl-4-pyrazolo[1,5-b]pyridazin-3-ylpyrimidin-2-amine series. The binding modes of compound $a1$ and compound $b1$ in the active sites of GSK3$\beta$, with Protein Data Bank (PDB) ID: 3L1S, are predicted by AutoDock Vina \cite{trott2010autodock} and visualized by PyMOL~\cite{delano2002pymol}. The key hydrogen bonds between compounds and the active sites are highlighted by yellow dash lines, while the green dash line refers to the intra-molecular halogen bond. The color intensity on each atom indicates its respective contribution to the prediction of our method, computed from a GNNExplainer \cite{gnnexplainer}.}
    \label{fig:ac_analysis}
\end{figure}

As shown in Figure~\ref{fig:ac_analysis}, visualized by PyMOL~\cite{delano2002pymol}, two series of compounds sharing the same scaffolds are potential inhibitors of glycogen synthase kinase-3 beta (GSK3$\beta$). The molecule activity cliff pairs, determined by the formulation in \cite{MoleculeACE}, are compounds <$a1, a2$>, <$a1, a3$>, <$b1, b2$>, and <$b1, b3$>. The explanations of both our and MolCLR's predictions are compared. In Figure~\ref{fig:ac_analysis}a, the potential intra-molecular halogen-bonding contact between the chlorine atom and hydrazone in compound $a1$, as indicated by the green dashed line, is disfavored for the inter-molecular hydrogen-bonding contact between the backbone carbonyl of the active site VAL-135 and hydrazone of the compound \cite{arnost20103}. As shown by our model explanation, compared to the compound $a2$ and $a3$, the chlorine atom of compound $a1$, along with its associated benzene ring, contribute less to the overall binding affinity prediction. It aligns well with the predicted binding mode. 

The predicted binding mode in Figure~\ref{fig:ac_analysis}b shows that the orientation of the substituent (i.e., the alkoxy group in compound $b1$ at the 6 position of pyrazolo[1,5-b]pyridazine) will cause steric clash with PHE-67 in the G-rich loop of GSK3$\beta$, thereby leading to a loss of potency~\cite{tavares2004n, stevens2008synthesis}. In contrast, the alkoxy groups at the 3' and 5' positions have no clear contact with the protein pocket. Remarkably, using only the topological information, our model can capture the importance of the influential alkoxy group at 6 position. In addition, the absence of the alkoxy group at 5' position (from compound $b2$ to $b3$) does not affect the overall atoms' contribution to the predicted potency. MolCLR performs equally good in terms of compound $b2$ and $b3$, but it fails to capture the influential alkoxy group in compound $b1$.

\section*{Discussion}
In this work, we propose a multi-channel learning framework for molecular representation learning, which aims to encode and utilize robust chemical knowledge that is generalizable to diverse downstream applications. Each channel is dedicated to learning unique self-supervised tasks, focusing on distinct yet correlated global and local aspects of the molecule. Specifically, MCD captures the molecule's global similarity, SCD highlights the fundamental role of the scaffold in affecting molecular characteristics, and CP targets the composition of functional groups. During fine-tuning, the model is able to identify which channel-wise representation is most relevant to the current application, thereby making the composite representation context-dependent.

One limitation of this framework is the need for a more effective prompt weight optimization mechanism. The initialization of prompt weights using roughness index can lead to sub-optimal performance. Since roughness index is a global QSPR metric that targets the overall chemical space, it does not account for any distribution shift between training and testing sets. This is the same for the other QSPR measures as well (e.g., SALI \cite{sali}, SARI \cite{sari}). As a result, the final representation performance may be less correlated with the initial roughness value under designated splits. This also explains the performance gap between Figure~\ref{fig:moleculeace}a and Figure~\ref{fig:moleculeace}b.

There are several interesting directions for future research. One promising direction is to incorporate different input representations into the framework. By merely leveraging topological molecular structure, the model is unable to differentiate molecular components with different conformations (e.g., functional groups' orientation or atom's chirality), which could significantly alter biochemical behaviors (see Supplementary Note 9 for further discussion). Besides, there exist other advanced data-driven techniques for studying the structural-activity relationship (SAR) that might be compatible with our framework. For example, Molecular Anatomy \cite{molecule_anatomy} argues that the network clustering from scaffold fragmentation and abstraction allows high quality SAR analysis. Such investigations aim to transfer knowledge from cheminformatics to machine learning models, potentially improving both model interpretability and robustness. More importantly, while our method has immediate implications for drug discovery, its molecular representation robustness further shed lights on its promising potential in other sub-fields of chemistry, such as materials science and environmental chemistry.

\section*{Methods} 

\paragraph{Graph neural networks (GNNs)} A graph $G=(V,E)$ is defined by a set of nodes $V$ and edges $E$. In the molecular graph, each node denotes an atom, and the edge denotes the chemical bond. Let $\mathbf{h_v}$ be the representation of node $v$, and $\mathbf{h_G}$ be the representation of the graph $G$. Modern GNNs follow the message-passing framework, such that node representations are updated iteratively via neighborhood aggregation:
\begin{align}
    \mathbf{h}_v^k=\operatorname{UPDATE}\left(\mathbf{h}_v^{k-1}, \operatorname{AGGREGATE}\left(\left\{\mathbf{h}_v^{k-1}, \mathbf{h}_u^{k-1}, e_{u v}\right\}: \forall u \in N(v)\right)\right) \text{ ,}
\end{align}
where $N(v)$ is the neighborhood of node $v$, $k$ denotes the layer index in a multi-layer GNN structure, and $e_{uv}$ denotes the edge connecting two nodes $u$ and $v$. The initialization of $\mathbf{h}_v^0$ comes from the predefined node features $\mathbf{x}_v$. The aggregate function integrates neighborhood information into the current node representation. The update function takes the updated node representation and the node representation from the previous $k-1$ layer and performs operations like concatenation or summation. After the iterative updates, a permutation-invariant pooling operation is performed to get the representation for the entire graph $G$: $\mathbf{h}_g = \operatorname{READOUT}(\mathbf{h}_v^k | v \in V)$. There are various number of options for the readout operation, including simple operations of mean and max, and advanced differentiable approaches like DiffPool \cite{diffpool} and GMT \cite{gmt}. 

{\paragraph{Roughness index} To analyze the structure-property relationship (SPR) within a chemical space, various quantitative metrics are proposed (e.g., SALI \cite{sali}, SARI \cite{sari}, ROGI \cite{roughness}). Even though the formulations are different, they all aim to capture the relationship between the representation difference and the property difference, which is also known as the molecular property landscapes. In this work, we mainly rely on ROGI value for both model training and experimental analysis. It is computed by first clustering the chemical space with different distance threshold $t$. For each cluster assignment, the weighted standard deviation $\sigma_t$ is calculated over the property labels of each cluster prototype. As the distance threshold increases, $\sigma_t$ will decrease monotonically from its initial value to zero (i.e., single cluster). If there are more similar molecules with similar labels, $\sigma_t$ will decrease slowly, and vice versa. Eventually, roughness index is formulated as below:
\begin{equation}
    \text{ROGI} = \int_{0}^{1} 2(\sigma_0 - \sigma_t)  \,dt
\end{equation}
The ROGI value has been shown to be strongly correlated with other metrics (e.g., SARI), as well as the representation performance on the given dataset \cite{roughness}. One of the main advantages of using ROGI is that it is a generalized QSPR metric applicable to a wide range of molecular representations, including both fingerprints and neural representations, while other metrics like SARI are tailored only for molecular fingerprints. This allows us to compute landscape roughness on pre-trained molecular representations (Supplementary Fig. 4 and Supplementary Note 3) and perform deeper QSPR analysis. }

\paragraph{Prompt-guided aggregation} Instead of using the same readout operations as the conventional GNNs, we adopt the prompt-guided aggregation, which is achieved using the multi-head attention. As shown in Figure~\ref{fig:framework}e, the embedding of the prompt token $\mathbf{h}_p$ is treated as the attention query, while the representations $\mathbf{h}_x$ of nodes/atoms $x$, are viewed as the keys and values. The resulted prompt-aware graph representation $\mathbf{h}_g^p = \sum_x \alpha_x \mathbf{v}_x$, where the attention weight for each node $x$ is computed as  $\alpha_x = \text{softmax}(\{\mathbf{q}\cdot \mathbf{k}_x / \sqrt{d_k}\}_x)$, and $\mathbf{q} = \mathbf{W}_q \mathbf{h}_p$, $\mathbf{k}_x = \mathbf{W}_k \mathbf{h}_x$, and $\mathbf{v}_x = \mathbf{W}_v \mathbf{h}_x$. Here $\{\mathbf{W}_q,\mathbf{W}_k,\mathbf{W}_v\}$ represent parameters in the linear transformations, and $\sqrt{d_k}$ is the scaling factor. Essentially, the prompt-aware $\mathbf{h}_g^p$ is aggregated from the weighted average of linear projection of $\mathbf{h}_x$. 

\paragraph{Adaptive margin contrastive loss} Contrastive learning is a technique widely used in self-supervised learning, which aims to group semantic similar samples closer while pushing dissimilar samples distant apart in the latent representation space. In this work, we adopt the triplet loss \cite{triplet} to formulate the contrastive learning. It considers triplets of data samples: the anchor $G_i$, the positive (i.e., semantic-similar) sample $G_i'$, and the negative (i.e., semantic-dissimilar) sample $G_j$. Its formulation is shown below.
\begin{align} \label{eq:triplet_vanilla} 
    \ell_{i, j(\neq i)} =\max\big(0, \alpha + d(\mathbf{h}_{g_i}, \mathbf{h}_{g_i}') - d(\mathbf{h}_{g_i}, \mathbf{h}_{g_j})\big) \text{ ,}
\end{align}
where $\mathbf{h}_{g_i}$ denotes the latent representation for sample $G_i$, and $\mathbf{h}_{g_i}'$ being the representation for its augmentation. The function $d(\cdot,\cdot)$ measures the L2 distance between two vectors. In general, this objective enforces the pair-wise distancing difference between <$G_i$, $G_i'$> and <$G_i$, $G_j$> to be at least $\alpha$ margin. However, as we discuss before, this formulation can only lead to a coarse-grained representation space, while neglecting the existing structural relationship among  molecules (e.g., shared rings or functional groups). Also, this formulation does not pose any constraints on the representation space beyond the margin (Supplementary Fig. 3). Therefore, we propose to add an additional contrastive formulation with adaptive margin among negative samples and the anchor.
\begin{align} \label{eq:triplet_vanilla_2} 
    \ell_{i, j(\neq i), k(\neq j \neq i)}  = & \max\big(0, \alpha_1(G_i, G_j) + d(\mathbf{h}_{g_i}, \mathbf{h}_{g_i}') - d(\mathbf{h}_{g_i}, \mathbf{h}_{g_j})\big) + \notag \\
    & \max\big(0, \alpha_1({G_i}, {G_k}) + d(\mathbf{h}_{g_i}, \mathbf{h}_{g_i}') - d(\mathbf{h}_{g_i}, \mathbf{h}_{g_k})\big) + \notag \\
    & \max\big(0, \alpha_2({G_i}, {G_j}, {G_k}) + 
    d(\mathbf{h}_{g_i}, \mathbf{h}_{g_j}) - d(\mathbf{h}_{g_i}, \mathbf{h}_{g_k})\big)
    \text{ ,}
\end{align}
where $\alpha_1(\cdot)$ and $\alpha_2(\cdot)$ are the adaptive margin functions. Let $\mathbf{z}_{g_i}$ be the conventional structural features of the sample $G_i$. We use the ECFP4 fingerprint \cite{morgan_fp}, which hashes circular atom neighborhoods into fixed-length binary strings, to represent the structural features. In molecule contrastive distancing, the adaptive function $\alpha_1(G_i, G_j) = \alpha_{\text{offset}} \times (1-\text{sim}(\mathbf{z}_{g_i}, \mathbf{z}_{g_j}))$, and $\alpha_2(G_i, G_j, G_k) = \alpha_{\text{offset}} \times (\text{sim}(\mathbf{z}_{g_i}, \mathbf{z}_{g_j}) - \text{sim}(\mathbf{z}_{g_i}, \mathbf{z}_{g_k}))$, where $\text{sim}(\cdot,\cdot)$ denotes the Tanimoto similarity. The scaffold contrastive distancing has the same formulation, except that the molecule sample $G$ is replaced by its scaffold $s(G)$. Note that the formulation now considers quadruplet of data samples <$G_i$, $G_i'$, $G_{j(\neq i)}$, $G_{k(\neq j \neq i)}$>. Even though the theoretical complexity is increased from $O(N^2K)$ to $O(N^3)$, we can perform fixed-size random sampling with respect to the computed similarity differences in $\alpha_2(\cdot)$. Quadruplets are also dropped if the computed values are negative. Eventually, the optimization goal is to minimize the loss summation as $\min \mathcal{L}_{adaptive} = \min \sum_{i,j,k} \ell_{i, j(\neq i), k(\neq j \neq i)}$.

{\paragraph{Regularization} To further improve the robustness of the pre-trained model, we incorporate two regularization schemes for the intra-channel node aggregation and the inter-channel alignment. The former strategy aims to ensure that all atoms contribute to MCD, while only the scaffold atoms contribute to SCD. This is accomplished by a smooth L1 loss between the attention score and the atom importance matrix. We realize that without any regularization, the aggregation module may rely on specific structural patterns to perform the SSL tasks, causing the attention distribution to skew towards certain substructures. This is also known as the shortcut learning \cite{shortcut} in deep learning models. As a result, it is possible to feed an incomplete view of the molecule to the property prediction layer (i.e., atoms that receive low attention across all channels) without the attention regularization. The latter regularization encourages better alignment of representation spaces. Since the three channels are learned separately via different tasks, the numeric values of representations at a given position may not be aligned. It means that the linear combination of representations before any fine-tuning may not be meaningful. To encourage the composite representation space to be better defined (e.g., subtracting $\mathbf{h}_g^{[MCD]}$ with $\mathbf{h}_g^{[SCD]}$ could represent the fragments beyond scaffold), we use a set of supervised tasks (e.g., predicting molecular weight and logP), along with the corresponding prompt weight presets, to regularize the channel alignment. These tasks mainly predict the molecule/scaffold descriptors using the composite representation. For example, the prompt weight preset for predicting the molecular weight would be [0.45, 0.1, 0.45], while the preset for predicting the weight of the scaffold would be [0.1, 0.45, 0.45]. We multiply all the regularization losses by a factor of 0.1 to prevent the regularization from dominating the learned representation.}

\paragraph{Prompt-guided multi-channel learning} The overall multi-channel learning framework is inspired by the work in \cite{wang2023multilevelprotein}. At the pre-train stage, the MCD and SCD channels perform the contrastive learning using the adaptive margin loss. Subgraph masking is used to generate positive samples for MCD. A subgraph is defined by a central atom along with its one-hop neighbors. The masking is performed at the attribute level, ensuring that the topological structure is retained. Meanwhile, the CP channel learns masked subgraph prediction as a multi-label classification task and motif prediction as a regression task: $\mathcal{L}^\text{CP} = \mathcal{L}_\text{CE} + \mathcal{L}_\text{SmoothL1}^{fg}$, where CE stands for cross-entropy loss, SmoothL1 for smooth L1 loss, and $fg$ for normalized functional group descriptors. Overall, the framework is optimized using the three channel losses along with the regularization losses: 
\begin{equation}
    \mathcal{L}_{overall} = \mathcal{L}_{adaptive}^\text{MCD} + \mathcal{L}_{adaptive}^\text{SCD} + \mathcal{L}^\text{CP} + 0.1 \times \mathcal{L}_{regu}
\end{equation}

At the finetune stage, the pre-trained model parameter are used to initialize the model. Besides, we introduce an additional prompt selection module to combine representations from different channels into the task-specific (i.e., context-dependent) composite representation. Essentially, it learns the relevance between different pre-trained molecular knowledge and the downstream application, hence bridging the gap between pre-training and fine-tuning objectives. To incorporate task-specific SPR information into the model at an early stage of fine-tuning, we propose to initialize the prompt weights from computing the roughness index (ROGI) \cite{roughness}. In short, the initial prompt weights should lead to a composite representation with the lowest ROGI value (i.e., smoothest quantitative structure-property landscape) with respect to the current application. A low ROGI value indicates smoother landscape, hence better modellability. We use a simple Bayesian optimization pipeline to find the best initialization with the lowest ROGI value on the training set. Essentially, the input parameters of the Bayesian optimization are $k-1$ learnable scalers, where $k$ is the number of channels. The black-box utility function first computes the distribution over $k$ values by treating the input scalars as logits, and calculates the composite representation as well as its corresponding roughness index. We utilize the Quasi MC-based batch Expected Improvement as the acquisition function. After the initialization, the entire model except for the prompt-guided aggregation module is optimized towards the molecular property labels. The parameters in the aggregation modules are fixed during fine-tuning as we aim to use it as a pooling layer independent of the downstream applications. As shown in Supplementary Figs. 7-9, the node attention scores within the aggregation module reflect the expected atom importance. For example, the SCD's aggregation attention captures mostly the scaffold atoms. We further discover that the learned prompt weights exhibit patterns that are well-aligned with the necessary chemical knowledge for solving the downstream tasks (Supplementary Fig. 12, Supplementary Table 1, and Supplementary Note 8).

\paragraph{Experimental setup} We pre-train our framework using the molecules from ZINC15 \cite{zinc}, which is an open-sourced database that contains 2 million unlabelled drug-like compounds. We use CReM \cite{crem}, an open-sourced molecule mutation framework, for the scaffold-invariant molecule perturbation. To be more specific, we first use RDKit \cite{rdkit} to identify the Bemis-Murcko scaffold of molecules, as well as the connection sites (i.e., atom indices) between the scaffold and the side chains. The CReM algorithm then takes these indices and performs fragment replacement using an external fragment pool. Perturbation results are further filtered by chemical validity and the maximum number of changed atoms allowed. At last, we are able to successfully compute perturbations for 1,882,537 out of 2 million molecules. These molecules form our final pre-train database. In the contrastive-based pre-training, we consider five positive samples for each anchor molecule. In molecule distancing, we randomly apply subgraph masking five times on the original molecule. In scaffold distancing, we randomly sample five scaffold-invariant perturbations. In terms of the model architecture, we follow the same architecture setup as the graph neural network in \cite{graphlog, gnnpretrain1, molclr, dsla}, which is a 5-layer Graph Isomorphism Network (GIN) \cite{gin} with hidden dimension size equals to 300. In addition, we also pre-train a GPS Graph Transformer model \cite{gps} with the same number of layer and hidden dimension size. We use the same molecule feature sets as works in \cite{gnnpretrain_grover, kano}. During fine-tuning, we utilize the scaffold splits method \cite{scaffoldsplit} for MoleculeNet \cite{MoleculeNet}, with a train, validation, and test ratio of 8:1:1. On the other hand, we apply the stratified splits for MoleculeACE \cite{MoleculeACE}, as proposed by itself, with a train, validation, and test ratio of 8:1:1. 

\paragraph{Baselines} We consider twelve baseline pre-training methods in our experiments. 1. GraphLoG \cite{graphlog} achieves a hierarchical prototypical embedding space with the conventional graph perturbation techniques; 2. D-SLA \cite{dsla} proposes the discrepancy learning to refine the embedding space with the conventional graph perturbation techniques; 3. MolCLR \cite{molclr} utilizes the NT-Xent \cite{simclr} contrastive loss with molecule perturbation techniques, including atom/bond editing and subgraph masking; 4. KANO \cite{kano} introduces the knowledge graph prompting techniques to augment molecular graph, while the augmentations are also learned in the contrastive manner. {5. MoLFormer \cite{molformer} performs large scale of masked language modeling on SMILES sequence using a linear attention Transformer.} 6. Hu et. al \cite{gnnpretrain1} adopts masked attribute prediction and molecular subgraph prediction; {7. KPGT \cite{kpgt} proposes the Line Graph Transformer on molecular graphs and performs knowledge prediction. It takes a masked molecular graph, molecule fingerprint, and molecule descriptors as input, and aims to reconstruct the molecule fingerprint and descriptors during pre-training.} 8. GROVER \cite{gnnpretrain_grover} proposes the GTransformer architecture and applies context prediction learning on molecular graphs; 9. GEM \cite{gem} encodes the 3D geometric information of molecules and predicts the bond angle and atom distance. {10. Uni-Mol \cite{uni-mol} also incorporates 3D information of the molecule and pre-trains a SE(3) Transformer \cite{se3transformer} using 3D position recovery and masked atom prediction. We further include 11. GraphMVP \cite{graphmvp}, which performs both contrastive and predictive learning between 2D toplogical and 3D geometric information of the molecule.} {12. ImageMol \cite{imagemol}, which leverages visual information of molecule images for molecular property prediction.} We also train the GIN \cite{gin} and GPS \cite{gps} models from scratch for comparison. We reproduce the performance of MolCLR, GROVER, GEM, KPGT, MoLFormer, and KANO on the MoleculeNet benchmark using their provided checkpoints and code repositories. Note that the reported results of GROVER, KPGT, MoLFormer and KANO in their original papers are evaluated under the balanced scaffold splits, which is different than the deterministic scaffold splits used in this work and the rest of the baseline methods. {Both splits aim to hold out a set of molecules with scaffolds that are not seen during training. The deterministic scaffold splits prioritizes the satisfaction of the specified split ratio, while the balanced scaffold splits prioritizes the balance of scaffold frequency across the training, validation, and test sets. We use the deterministic scaffold splits in our experiments primarily because most of the baselines also employ it. All other aspects, including hyperparameter choices for both the models and the training setup, remain consistent with the default configurations of these baseline methods for replicating the results. For the MoleculeACE benchmark, we leverage their corresponding repositories, default configurations, and the provided model checkpoints of these baseline methods and fine-tune each task using either MoleculeACE's stratified splits or random splits.}

\paragraph{Representation space probing} We propose to analyze the dynamics of representation space during fine-tuning to evaluate the representation robustness and generalizability. To be more specific, we probe the mapping of the representation space at five different training timestamps (epoch 0, epoch 10, epoch 20, epoch 50, epoch 100). The 2-D view is constructed via the T-SNE \cite{tsne} dimension reduction technique. Besides the visualization, we also report four additional metrics that capture the representation characteristics. We report the ROGI value and Rand index in the training set, plus the R-squared value and cliff-noncliff distance ratio in the validation set. The Rand index is calculated from two clustering assignments. The first clustering is formed using k-means clustering with the molecule ECFP4 fingerprint (radius=2, nBits=512). The second clustering is formed using k-means clustering with the representation at the current timestamp. We set the same number of clusters for these two clusterings. 
In terms of the cliff-noncliff distance ratio, we first identify the cliff and the noncliff molecule pairs using the same principle in MoleculeACE \cite{MoleculeACE}. Then, we compute and average the representation distance of each pair. At last, the cliff-noncliff ratio is formed by normalizing their average distance.

\bgroup
\def\arraystretch{1.4}%
\begin{table}[]
\caption{Fine-tuning results on 7 classification tasks in MoleculeNet {using the scaffold splits}. Average Receiver Operating Characteristic Area Under the Curve (ROC-AUC) value is reported, along with the score standard deviation (shown by $\pm$) from three independent runs. The first two rows of GIN and GPS showcase the performance of the backbone models without pre-training. Best performance is shown in bold.}
\label{tab:moleculenet}
\centering
\resizebox{\textwidth}{!}{
\begin{tabular}{lccccccc|c}
\Xhline{1.5pt} 
\textbf{Methods}   & BBBP        & Clintox     & MUV         & HIV         & BACE        & Tox21       & SIDER    & Avg.  \\
\textbf{\#task}  & 1 & 2 & 17 & 1 & 1 & 12 & 27  & \\ \hline
GIN \cite{gin}          & 65.8 $\pm$ 4.5 & 58.0 $\pm$ 4.4 & 71.8 $\pm$ 2.5 & 75.3 $\pm$ 1.9 & 70.1 $\pm$ 5.4 & 74.0 $\pm$ 0.8 & 57.3 $\pm$ 1.6  &  67.5 \\
GPS \cite{gps}          & 64.8 $\pm$ 3.0 & 87.2 $\pm$ 0.9 & 69.8 $\pm$ 3.8 & 73.1 $\pm$ 3.3 & 78.0 $\pm$ 3.0 & 74.5 $\pm$ 0.6 & 60.8 $\pm$ 0.6  &  72.6 \\ \hline
Hu et. al \cite{gnnpretrain1}           & 70.8 $\pm$ 1.5 & 72.6 $\pm$ 1.5 & 81.3 $\pm$ 2.1 & {79.9 $\pm$ 0.7} & 84.5 $\pm$ 0.7 & {78.7 $\pm$ 0.4} & 62.7 $\pm$ 0.8   &   75.8\\
GraphLoG \cite{graphlog}            & 72.3 $\pm$ 0.9 & 74.7 $\pm$ 2.2 & 74.2 $\pm$ 1.8 & 75.4 $\pm$ 0.6 & 82.2 $\pm$ 0.9 & 75.1 $\pm$ 0.7 & 61.2 $\pm$ 1.1  & 73.6\\
D-SLA \cite{dsla}               & 72.6 $\pm$ 0.8 & 80.2 $\pm$ 1.5 & 76.6 $\pm$ 0.9 & 78.6 $\pm$ 0.4 & 83.8 $\pm$ 1.0 & 76.8 $\pm$ 0.5 & 60.2 $\pm$ 1.1 & 75.5\\
MolCLR \cite{molclr}             & 73.5 $\pm$ 0.4 & 90.4 $\pm$ 1.7 & 75.5 $\pm$ 1.8 & 77.6 $\pm$ 3.2 & 83.5 $\pm$ 1.8 & 76.7 $\pm$ 2.1 & 60.7 $\pm$ 5.7 & 76.8\\
GraphMVP \cite{graphmvp}             & 72.4 $\pm$ 1.6 & 79.1 $\pm$ 2.8 & 77.7 $\pm$ 0.6 & 77.0 $\pm$ 1.2 & 81.2 $\pm$ 0.9 & 75.9 $\pm$ 0.5 & 63.9 $\pm$ 1.2 & 75.3\\
GROVER \cite{gnnpretrain_grover}              & 69.5 $\pm$ 0.1 & 76.2 $\pm$ 3.7 & 67.3 $\pm$ 1.8 & 68.2 $\pm$ 1.1 & 81.0 $\pm$ 1.4 & 73.5 $\pm$ 0.1 & 65.4 $\pm$ 0.1 & 71.6\\
GEM \cite{gem}                & 71.8 $\pm$ 0.6 & 89.7 $\pm$ 2.0 & 77.0 $\pm$ 1.5 & 78.0 $\pm$ 1.4 & 84.9 $\pm$ 1.1 & 78.2 $\pm$ 0.3 & {67.2 $\pm$ 0.6} & 78.1 \\
ImageMol \cite{imagemol}  & 73.9 $\pm$ 0.2 & 85.1 $\pm$ 1.4 & \textbf{82.5} $\pm$ 0.8 & 79.7 $\pm$ 0.2 & 83.9 $\pm$ 0.5 & 77.3 $\pm$ 0.1 & 66.0 $\pm$ 0.1  & 78.3\\
KANO \cite{kano}                & 69.9 $\pm$ 1.9 & 90.7 $\pm$ 2.2 & 74.7 $\pm$ 2.0 & 75.7 $\pm$ 0.3 & 82.7 $\pm$ 0.9 &  75.8 $\pm$ 0.5  &  60.2 $\pm$ 1.4 & 75.7 \\
KPGT \cite{kpgt}                 & 71.4 $\pm$ 0.7 & 88.8 $\pm$ 2.9 & 75.7 $\pm$ 1.4 & 77.9 $\pm$ 1.2 & 81.8 $\pm$ 2.7 &  78.5 $\pm$ 0.5  &  64.7 $\pm$ 1.0 & 77.0 \\
MoLFormer \cite{molformer}       & 70.9 $\pm$ 1.0 & 91.1 $\pm$ 0.9 & 80.5 $\pm$ 1.5  & 76.7 $\pm$ 0.4 & 83.6 $\pm$ 1.1 &  77.3 $\pm$ 0.4  &  64.9 $\pm$ 0.7 & 77.8  \\
Uni-Mol \cite{uni-mol}       & 72.9 $\pm$ 0.6 & 91.9 $\pm$ 1.8 & 82.1 $\pm$ 1.3 & \textbf{80.8 $\pm$ 0.3} & 85.7 $\pm$ 0.2 &  \textbf{79.6 $\pm$ 0.5}  &  65.9 $\pm$ 1.3 & 79.8 \\
\hline
Ours$_\text{GIN}$                 & \textbf{74.1 $\pm$ 0.6} & \textbf{95.7 $\pm$ 1.2} & 81.2 $\pm$ 0.5 & 79.8 $\pm$ 0.3 & 85.0 $\pm$ 1.1 & 77.5 $\pm$ 0.3 & 66.7 $\pm$ 0.8 & {80.1} \\ 
Ours$_\text{GPS}$                 & {73.6 $\pm$ 0.7} & {95.1 $\pm$ 0.5} & 81.5 $\pm$ 0.8 & 80.2 $\pm$ 0.5 & \textbf{86.1 $\pm$ 1.3} & 79.0 $\pm$ 0.6 & \textbf{68.7 $\pm$ 0.2} & \textbf{80.6} \\
\Xhline{1.5pt} 
\end{tabular}}
\end{table}
\egroup

\section*{Data Availability}
All relevant data supporting the key findings of this study are publicly available. All datasets utilized in this paper can be found at \url{https://github.com/yuewan2/MolMCL} or the Zenodo repository at \url{https://doi.org/10.5281/zenodo.14010436} \cite{myrepo}. This includes the processed pre-train dataset of ZINC15, the MoleculeNet benchmark (originally from \url{https://github.com/deepchem/deepchem}) and the MoleculeACE benchmark (originally from \url{https://github.com/molML/MoleculeACE}). The reference protein structure for 3L1S used in this study is available in the Protein Data Bank under accession code 3L1S [\url{https://www.rcsb.org/structure/3l1s}]. Source data are provided with this paper. 

\section*{Code Availability}
The source code of this work can be accessed via the GitHub repository at \url{https://github.com/yuewan2/MolMCL} or the Zenodo repository at \url{https://doi.org/10.5281/zenodo.14010436} \cite{myrepo}.

\medskip

\clearpage


\section*{Acknowledgements}
We have no specific acknowledgments to report for this work.

\section*{Author contributions}
Y.W. proposed and implemented the methodology, and conducted the computational experiments. J.W. carried out the chemical analysis of the model's predictions. T.H., C.H., and X.J. provided guidance throughout the project. Y.W. and J.W. led the paper writing and all authors contributed.

\section*{Computing Interests}
All authors declare no competing interests.

\clearpage

\end{document}


\flushbottom

\maketitle	

\section{Representation similarity}
\begin{figure}[!t]
    \centering
    \includegraphics[width=1\columnwidth, trim={0cm 20.8cm 0cm 0.5cm},clip]{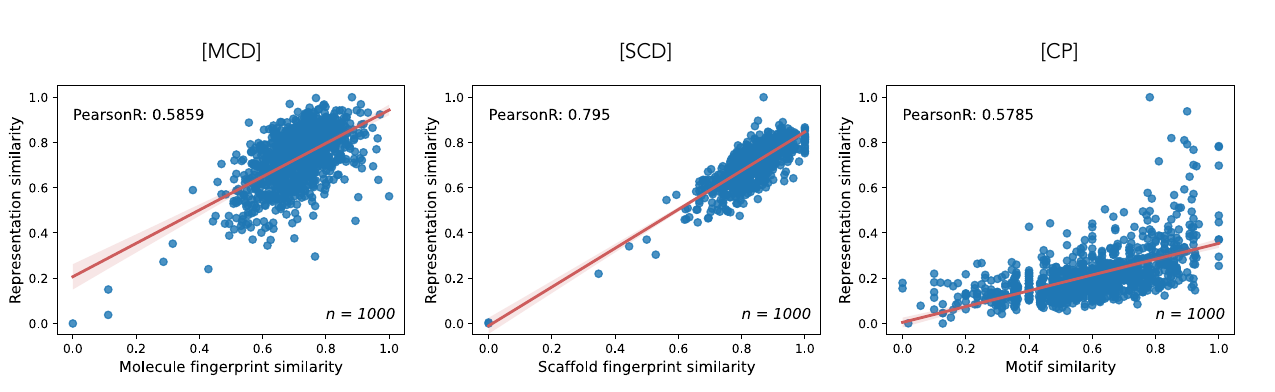}
    \caption{\textbf{Comparison of conventional similarity measures and representation distance.} For illustration purposes, 1000 molecule pairs are randomly sampled from ZINC15 \cite{zinc}. The x-axis represents the molecular similarity measured by Tanimoto similarity over structural features (i.e., fingerprint, scaffold fingerprint, functional group composition vector), while the y-axis represents the similarity measured by the normalized L2 norm over learned representations. Pearson correlation coefficient is reported.}
    \label{fig:correlation}
\end{figure}

We compare the relationship between the conventional fingerprint/descriptor similarity measures and the learned representation distances across channels (i.e., MCD, SCD, and CP), we randomly sample 1000 molecule pairs from ZINC15 \cite{zinc} and examine the correlation between the conventional molecule similarity measures and the learned representation distances. To be more specific, we compute the Tanimoto similarity between molecule ECFP4 fingerprints \cite{morgan_fp}, scaffold ECFP4 fingerprints, and functional groups (i.e., motif) descriptors \cite{gnnpretrain_grover}. We compute and normalize the L2 norm distance between the learned representations. As shown in Figure~\ref{fig:correlation}, the representation distance is highly correlated with the conventional similarity measures. This alignment also confirms that the representations learned from different channels indeed capture different global-local perspectives of the molecule.  

\section{Implicit representation hierarchy}
As previously discussed, the representations learned via different channels can elicit different levels of global and local molecule attributes. Molecule distancing focuses on the global view of the molecule, scaffold distancing targets the partial view, and context prediction tackles the local view (i.e., functional groups composition). To better visualize this representation hierarchy, we perform an iterative clustering algorithm over the molecule space of the CHEMBL237 dataset. Specifically, we conduct three stages of clustering with respect to the three channeled representations. Each stage considers the clustered results from the previous stage and further refines the clustered subspace. In hypothesis, representations derived from context prediction (i.e., $\mathbf{h}_g^{\text{[CP]}}$) may result in a coarse-grained clustering in terms of the intra-cluster molecule similarity. This is because molecules with identical functional groups can still elicit structural differences. In comparison, representations from scaffold distancing (i.e., $\mathbf{h}_g^{\text{[SCD]}}$) could yield a less coarse-grained clustering by grouping molecules with similar scaffolds together. Representations from molecule distancing (i.e., $\mathbf{h}_g^{\text{[MCD]}}$) consider structural variations beyond the scaffold, potentially leading to a more fine-grained clustering. 

\begin{figure}[!ht]
    \centering
    \includegraphics[width=1\columnwidth, trim={1.4cm 7cm 1.4cm 1.5cm},clip]{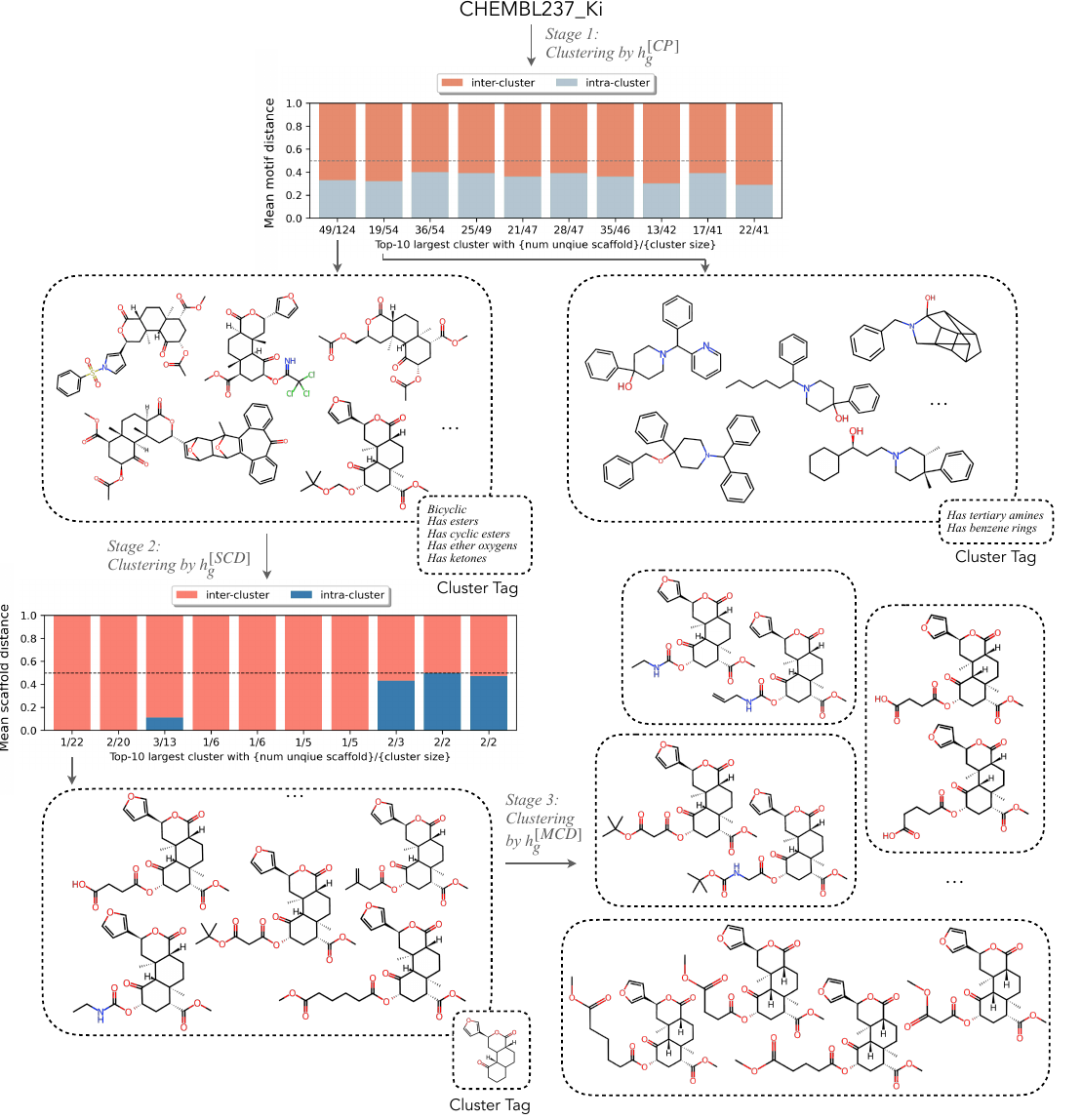}
    \caption{\textbf{Hierarchical clustering by the multi-channel representation.} The molecules in CHEMBL237\_Ki \cite{MoleculeACE} ($n=2602$) are initially clustered using the learned representation $\mathbf{h}_g^{[CP]}$, which emphasizes the local composition of each molecule. The bar chart shows the inter-cluster and intra-cluster difference of average motif distance, measured by 1 - Tanimoto similarity over the functional group composition vectors. Each cluster is then further divided into subsequent sub-clusters using $\mathbf{h}_g^{[SCD]}$ and $\mathbf{h}_g^{[MCD]}$. Scaffold distance is measured by the 1 - Tanimoto similarity over the scaffold fingerprint. The properties of each sub-cluster are indicated by their cluster tags and visualized through sample illustrations.}
    \label{fig:clustering}
\end{figure}

As shown in Figure~\ref{fig:clustering}, we first perform clustering with $\mathbf{h}_g^{\text{[CP]}}$. The bar chart illustrates the normalized ratio of intra-cluster motif distance to inter-cluster motif distance from the top-10 largest clusters. It suggests that intra-cluster molecules are more likely to have similar functional groups composition. Peaking inside the clusters, examples show the shared set of functional groups among the grouped molecules, which is denoted as the cluster tag. The two numbers at the x-axis represent the number of unique scaffolds and the cluster size, respectively. It means that the grouped molecules at this stage are not necessarily structurally similar. We further perform the stage two clustering using $\mathbf{h}_g^{\text{[SCP]}}$ on top of the stage 1 results. Again, the bar chart (with darker color) illustrates the normalized ratio of intra-cluster scaffold distance to inter-cluster scaffold distance. Five out of ten clusters contain molecules with same scaffolds. At stage three, the molecule space are further refined using representation $\mathbf{h}_g^{\text{[MCP]}}$. Despite the grouped molecules from stage two already share common scaffolds, the representation could still capture the minor structural differences in the terminal side chains as shown by the examples. From coarse-grained to fine-grained, the cluster becomes increasingly refined after iteratively applying clustering along the multi-channeled representations. It supports the hypothesis that representations from different channels can capture different views of the molecules from global to local.

\section{Roughness comparison}
\label{sec:si_roughness}

\begin{figure}[!t]
    \centering
    \includegraphics[width=1\columnwidth, trim={0.4cm 1.2cm 0cm 0cm},clip]{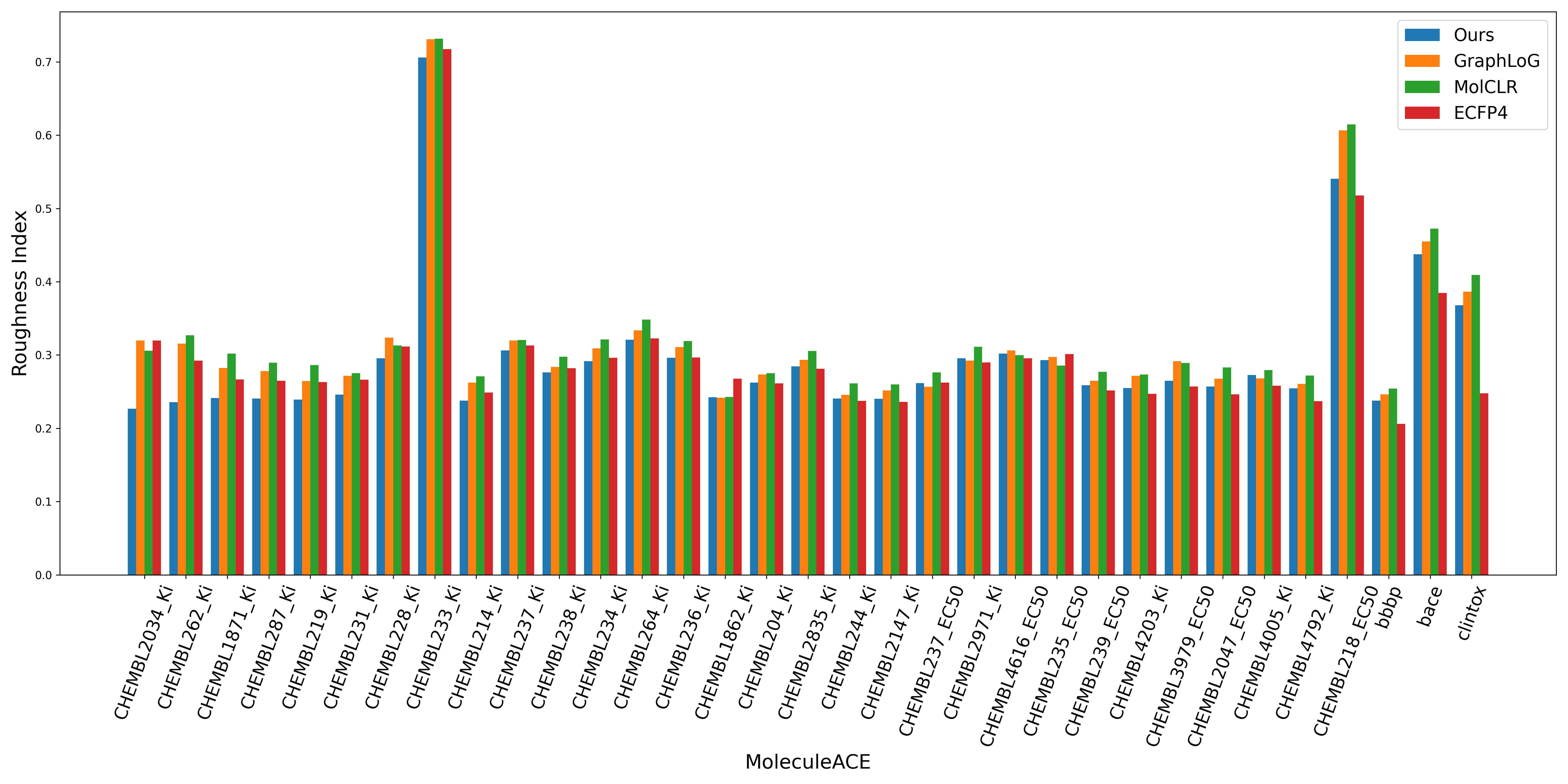}
    \caption{\textbf{Roughness index (ROGI) comparison.} The computed ROGI \cite{roughness} value of the representations (Ours$_\text{GIN}$, GraphLoG \cite{graphlog}, MolCLR \cite{molclr}, and ECFP4 fingerprint \cite{morgan_fp}) against molecular property measures in 33 datasets \cite{MoleculeACE, MoleculeNet}. Lower ROGI value indicates smoother molecular property landscape. The bars are arranged from left to right based on the descending order of the ROGI value differences between our representation and the most effective alternative representation.} 
    \label{fig:roughness}
\end{figure}

In this section, we report the roughness index (i.e., ROGI) \cite{roughness} difference between our method and two SSL baseline methods, GraphLoG and MolCLR, prior to fine-tuning on 33 datasets. We also consider the ECFP4 fingerprint \cite{morgan_fp} as one of the traditional molecule representations used in machine learning. The ROGI value captures the roughness of the property landscape given a representation. Low ROGI value indicates better modellability, and is strongly correlated with the convergence rate and the overall performance at the fine-tune stage. As shown in Figure~\ref{fig:roughness}, our representation has better ROGI value compared to other deep molecule representations in 22 out of 33 datasets. Its ROGI value is also lower than that of fingerprint in 18 out of 33 datasets. The advantage of the fingerprint representations is consistent with the finding in \cite{pretrainhelpornot}. However, it does not guarantee that the fine-tuning performance using fingerprints must outperform other deep learning methods. Roughness metric in general analyzes the overall chemical space, without considering any distribution shift between training and testing sets. It means that representations with smoother structure-property landscape may not necessarily perform better under data splits that introduce distribution shift (e.g., scaffold split in MoleculeNet \cite{MoleculeNet} and stratified split in MoleculeACE \cite{MoleculeACE}). Another reason is because the pre-trained molecule representations serve only as initialization for the embedding space. The model parameters, which are responsible for outputting these representations, also undergo parameter updates during the fine-tuning optimization process. In other words, the updated representations continue to benefit from the flexibility and strength of deep learning models.

\section{Convergence rate}
\begin{figure}[!t]
    \centering
    \includegraphics[width=1\columnwidth, trim={0.8cm 17.2cm 0.8cm 0.3cm},clip]{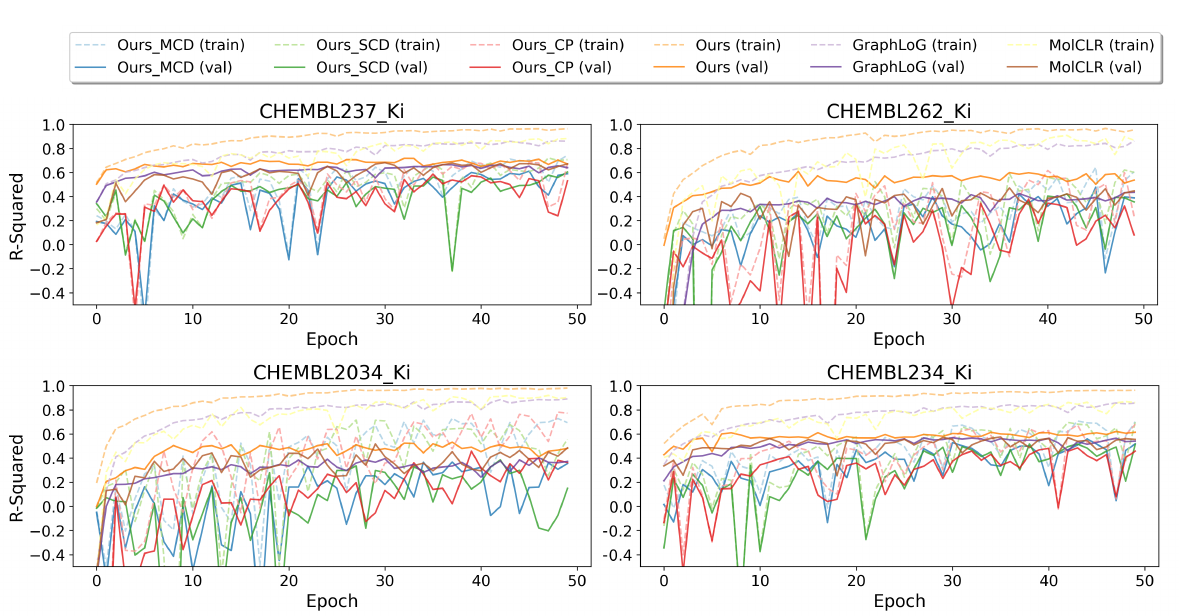}
    \caption{\textbf{Convergence rate at fine-tuning.} R-squared curves of Ours$_\text{GIN}$, GraphLoG \cite{graphlog}, MolCLR \cite{molclr} on both training and validation set of four datasets are ploted. Ours\_MCD, Ours\_SCD, and Ours\_CP represent the use of only the corresponding channel for fine-tuning. The dashed line represents the training R-squared value, while the solid line corresponds to the validation R-squared value.}
    \label{fig:convergence}
\end{figure}

We present a comparative analysis of the downstream convergence rates among our proposed method, GraphLoG, and MolCLR in four example datasets (CHEMBL237\_Ki, CHEMBL262\_Ki, CHEMBL2034\_Ki, and CHEMBL234\_Ki). As illustrated in the Figure~\ref{fig:convergence} , our representation facilitates faster convergence rate in terms of both the training and validation R-squared value. It suggests that our representation exhibits better downstream transferability, which is consistent with the conclusions on modellability discussed in Section~\ref{sec:si_roughness} in the main text. To further understand the efficacy of our composite representation achieved through prompt weight selection, we also consider the downstream performance when solely fine-tuned from individual channels, represented by Ours\_MCD, Ours\_SCD, and Ours\_CP. A notable observation is that their performance often exhibits high volatility, such that both the training and validation R-squared value oscillates significantly between epochs. One plausible explanation is that each corresponded prompt-guided aggregation module considers atoms with varying degrees of importance. In other words, individual aggregations may only tackle a partial view of the molecule, especially for the SCD and CP channels. Given that the aggregation modules remain fixed during fine-tuning, these partial perspectives may fall short in comprehensively learning the structure-property relationship.

\section{Representation robustness}
\begin{figure}[!t]
    \centering
    \includegraphics[width=1\columnwidth, trim={0.7cm 14.5cm 1.5cm 1.5cm},clip]{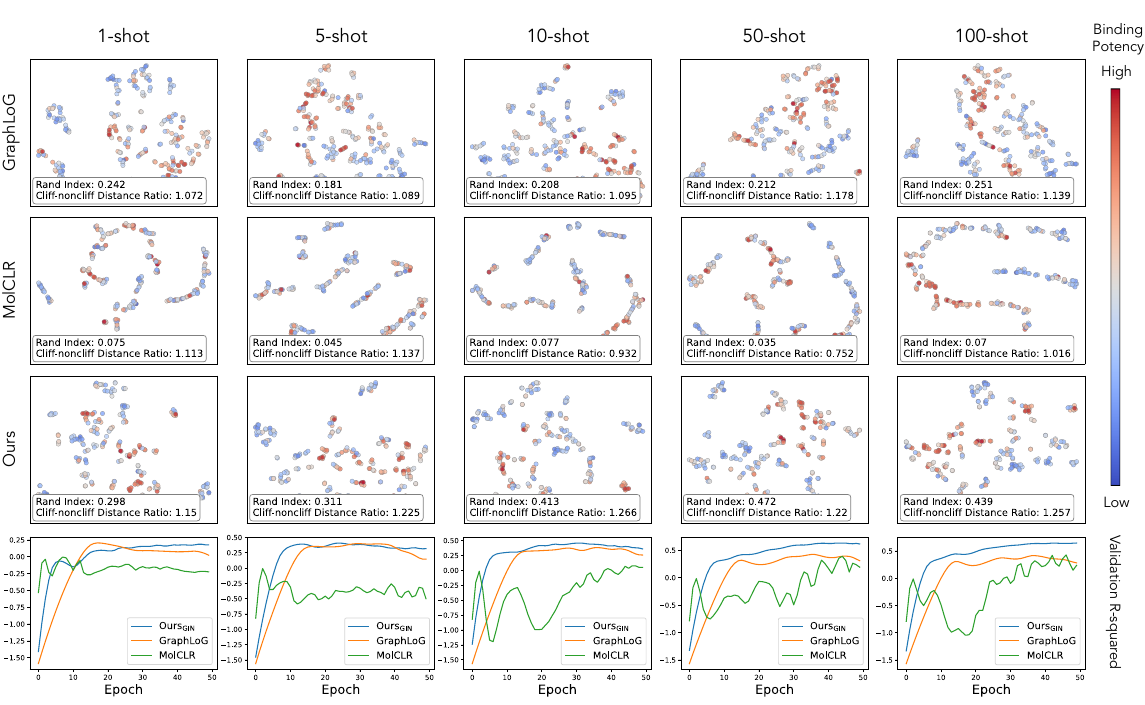}
    \caption{\textbf{Representation space probing (few-shot scenario).} The dynamics of molecule representation space of Ours$_\text{GIN}$, GraphLoG \cite{graphlog}, and MolCLR \cite{molclr} under five few-shot settings on dataset CHEMBL237\_Ki \cite{MoleculeACE} ($n=2602$). For each row, a 2D view of the representation space of the validation set is visualized. The dot coloring represents the normalized potency labels. Rand index \cite{rand_index} and cliff-noncliff distance ratio are reported, along with the validation R-squared value along the training process (bottom row).}
    \label{fig:fewshot_probing}
\end{figure}

In the main text, we evaluate the shift in representation space with three methods (Ours$_\text{GIN}$, GraphLoG \cite{graphlog}, and MolCLR \cite{molclr}) when fine-tuned on the CHEMBL237\_Ki dataset. In this section, we further analyze the representation performance under the few-shot scenarios. Specifically, we evaluate the fine-tuned results of the validation set using only 1\%, 5\%, 10\%, 50\%, and 100\% of the training data, corresponding to the 1-shot, 5-shot, 10-shot, 50-shot, 100-shot columns as shown in Figure~\ref{fig:fewshot_probing}. Each visualization shows what the representation space looks like after training for 100 epochs. Following the work in \cite{MoleculeACE}, instead of randomly sampling the training data, we perform stratified sampling by first applying k-means clustering using the molecule fingerprint, and selecting the required amount of data samples that are diverse across the clustering. We report the same Rand index \cite{rand_index} and cliff-noncliff distance ratio as the analysis in the main text. As shown in the figure, even with 1\% of training data, the representation space of MolCLR distorts significantly. The small rand index implies that large amount of structural information is lost, which can also be interpreted as label overfitting. GraphLoG, compared to MolCLR, shows a more balanced representation space mapping throughout all few-shot scenarios, also indicated by its consistently stable rand index around 0.24. Our method also maintains high Rand index across all few-shot scenarios. When comparing the cliff-noncliff distance ratio, our method consistently performs better than others with larger distance ratio. As shown by the validation R-squared curves, our method also achieves better performance in terms of the highest R-squared value, faster convergence rate, and training stability.

\section{Case study in prompt-guided node aggregation}
\begin{figure}[!t]
    \centering
    \includegraphics[width=1\columnwidth, trim={1cm 7cm 1cm 2cm},clip]{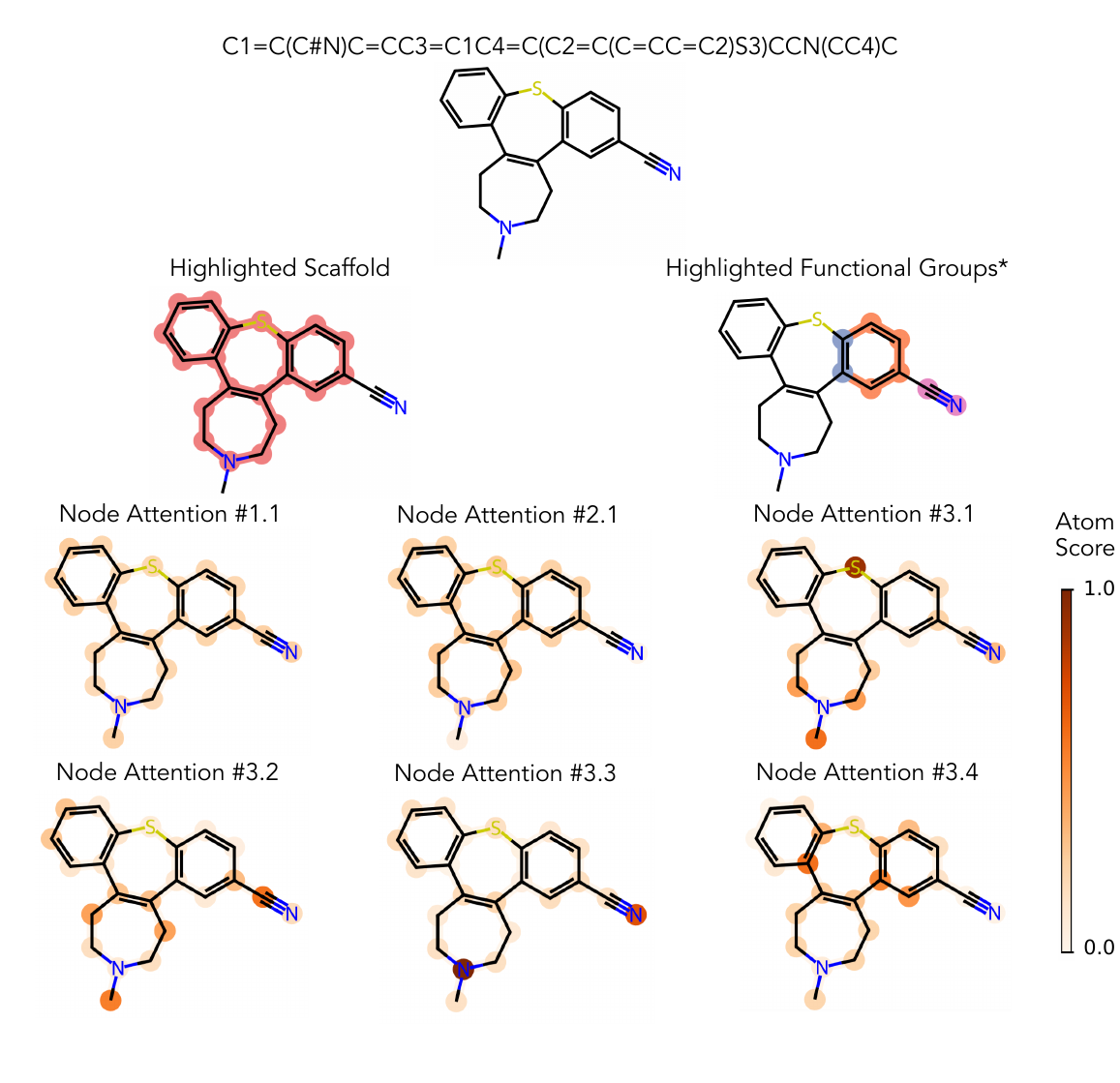}
    \caption{{\textbf{Case study of prompt-guided aggregation.} Visualization of atom aggregation attention on molecule 18-methyl-8-thia-18-azatetracyclo[13.5.0.02,7.09,14]icosa-1(15),2(7),3,5,9,11,13-heptaene-4-carbonitrile, along with its scaffold and functional groups. For illustration purposes, only certain functional groups are highlighted considering the atom overlaps (indicated by the *). The node attention id follows the format of \#$<$channel\_id$>$.$<$head\_id$>$. Six different attention distributions are shown from the prompt-guided aggregation. Darker color means higher atom importance.}}
    \label{fig:node_attn_1}
\end{figure}

\begin{figure}[!t]
    \centering
    \includegraphics[width=1\columnwidth, trim={0cm 11.5cm 0cm 2cm},clip]{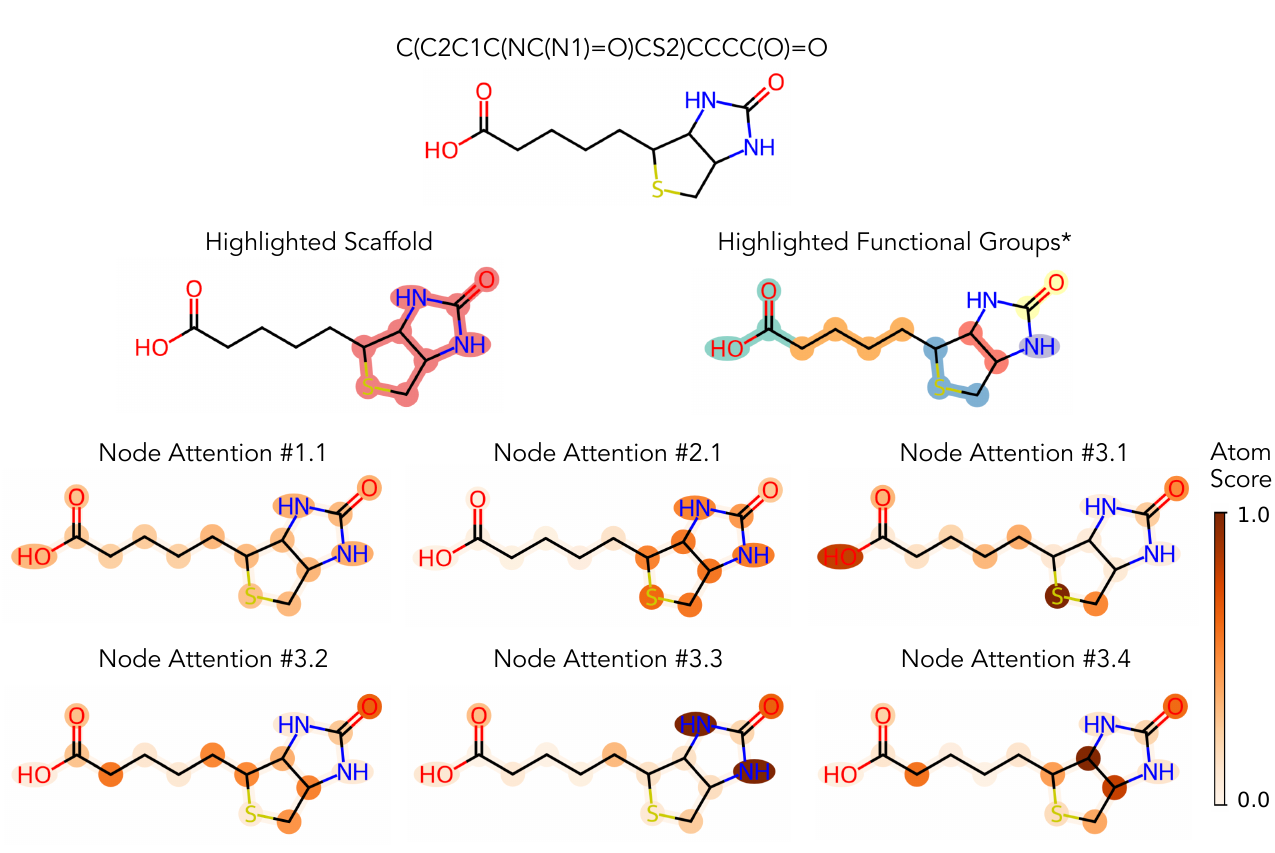}
    \caption{{\textbf{Case study of prompt-guided aggregation.} Visualization of atom aggregation attention on molecule 5-(2-oxo-1,3,3a,4,6,6a-hexahydrothieno[3,4-d]imidazol-4-yl)pentanoic acid, along with its scaffold and functional groups. For illustration purposes, only certain functional groups are highlighted considering the atom overlaps (indicated by the *). The node attention id follows the format of \#$<$channel\_id$>$.$<$head\_id$>$. Six different attention distributions are shown from the prompt-guided aggregation. Darker color means higher atom importance.}}
    \label{fig:node_attn_2}
\end{figure}

To interpret the prompt-guided aggregation, we randomly choose two out-of-distribution molecules from BBBP \cite{MoleculeNet} (beyond the pre-train datasets) and visualize their node attention scores, as shown in Figure~\ref{fig:node_attn_1} and Figure~\ref{fig:node_attn_2}. The highlighted scaffold corresponds to the Bemis-Murcko scaffold, while different functional groups are highlighted using different colors. For illustrative purposes, not all the functional groups are shown because of the atom overlaps. The node attention id corresponds to \#$<$channel\_id$>$.$<$head\_id$>$. For example, node attention \#1.1 is taken from the first attention head in the first channel (MCD). For the predicted node  attention, darker color indicates higher atom contribution, and vice versa. As illustrated by the figures, the node aggregation module is able to capture the expected atom importance. All atoms contribute to the molecular representation of the MCD channel, while only the atoms belonging to the scaffold are treated important for the SCD channel. When it comes to the node attentions computed for the CP channel, the attention scores of each head is a lot messier than before. One of the reasons is because CP is not only doing functional group composition prediction, but also masked subgraph prediction. Also, it is impossible to have a one-to-one mapping from attention heads to functional groups. However, we could still observe some interesting local patterns captured by the attention. For example, the node attention \#3.4 in Figure~\ref{fig:node_attn_2} seems to capture the bicyclic compound, the carbonyl group, as well as the thioether of the molecule. 

\begin{figure}[!t]
    \centering
    \includegraphics[width=1\columnwidth, trim={0.5cm 21.6cm 0cm 2cm},clip]{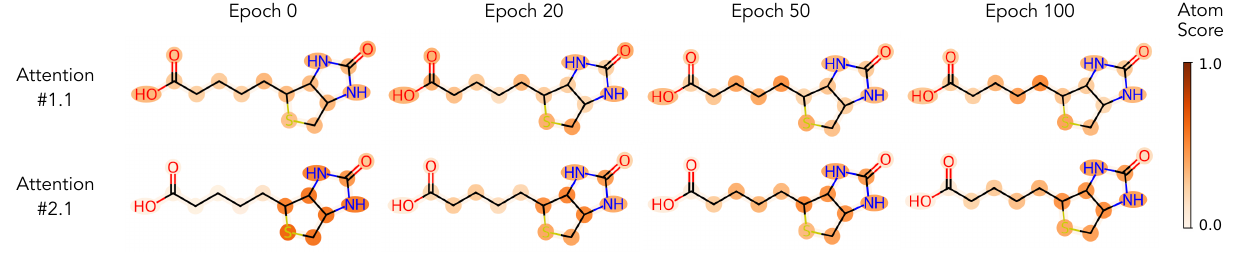}
    \caption{{\textbf{Case study of prompt-guided aggregation during fine-tuning.} Visualization of aggregation attention on molecule 5-(2-oxo-1,3,3a,4,6,6a-hexahydrothieno[3,4-d]imidazol-4-yl)pentanoic acid, when fine-tuning BBBP \cite{MoleculeNet} at epoch 20, 50, and 100. The node attention id follows the format of \#$<$channel\_id$>$.$<$head\_id$>$. Darker color means higher atom importance.}}
    \label{fig:node_attn_2_finetune}
\end{figure}

Even though the prompt-guided node aggregations are fixed during fine-tuning, the graph encoder (e.g, GNN) is still tunable. Therefore, we further investigate whether the aggregation patterns hold during fine-tuning, especially for the first two channels. Surprisingly, we realize that there is only a small shift in aggregation patterns. As shown in Figure~\ref{fig:node_attn_2_finetune}, aggregation patterns from the first two channels are visualized at epoch 20, 50, and 100 when fine-tuning the BBBP dataset. During fine-tuning, the score distribution of attention \#1.1 changes slightly, but it continues to span across all atoms. For attention \#2.1, the non-scaffold atoms, especially for the carbon chain in the example, begin to receive increased attention. However, their contributions to the aggregation remain lower than those of the scaffold atoms. We hypothesize that the fixed aggregation module acts as a constraint, limiting the degree of freedom of the graph encoder module during fine-tuning to ensure that the aggregation output is meaningful.

\section{Ablation study}

To gain a clearer understanding of the effects of each component on downstream performance, we conduct a comprehensive ablation study. Specifically, the ablation controls the main components in our pre-training design (i.e., adaptive margin loss, multi-channel learning framework, and regularizations) and fine-tuning strategies (i.e., simple concatenation of channel-wise representations versus learnable prompt selection module). Additionally, we also examine the relationship between individual channels and the chemical knowledge necessary for learning molecular properties.

\subsection{Pre-train components}

\begin{figure}[!t]
    \centering
    \includegraphics[width=1\columnwidth, trim={0cm 0cm 0cm 0cm},clip]{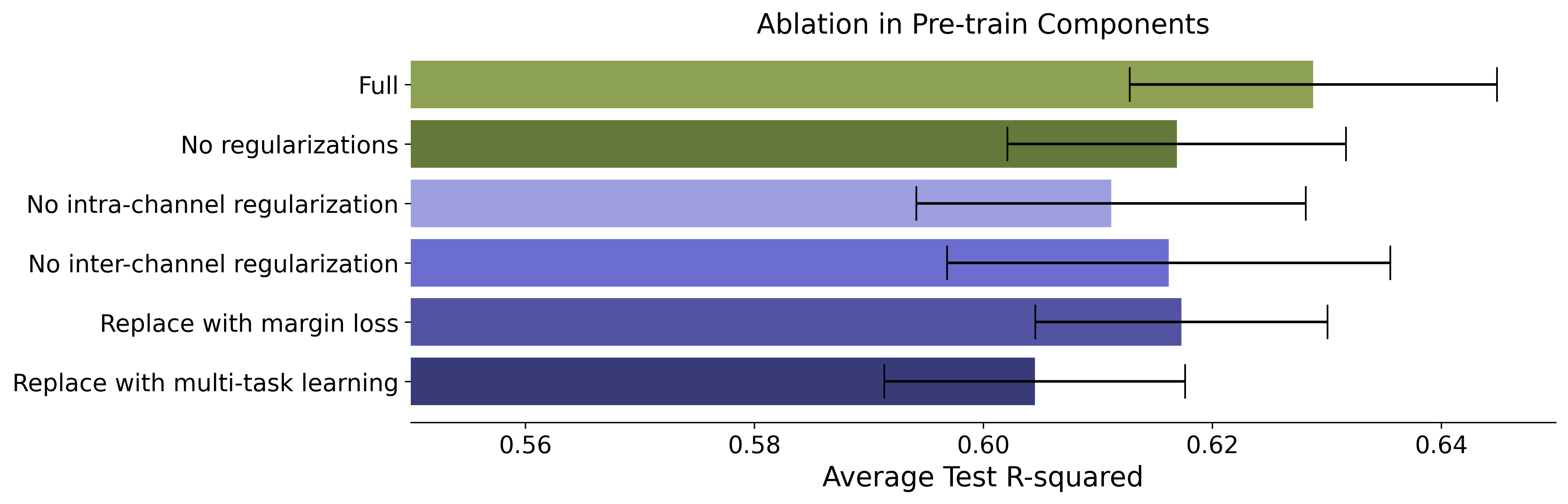}
    \caption{{\textbf{Ablation study in pre-train components.} The average performance on the 30 datasets in MoleculeACE \cite{MoleculeACE} under 6 pre-train settings. The error bars represent the average standard deviation of the standard deviation across the three independent runs for each dataset. See Table~\ref{tab:pretrain_ablation_detail} for more details.}}
    \label{fig:pretrain_ablation}
\end{figure}

We compare the model performance on the MoleculeACE benchmark across 6 different pre-training settings: 1. The full pre-training setting, corresponding to the best model configuration described in the main text. 2. Without any regularizations. 3. Without intra-channel regularization on node aggregation patterns. 4. Without inter-channel regularization on channel alignment. 5. Replacing the adaptive margin loss with the conventional margin loss. 6. Replacing the multi-channel learning with conventional multi-task learning, where there is only a single aggregation channel that learns all the pre-trained tasks. All settings, including the full setting, are pre-trained using GIN \cite{gin} as the model backbone for 40 epochs. Figure~\ref{fig:pretrain_ablation} presents the performance comparison across the 30 MoleculeACE datasets, measured by the average test R-squared value. As before, performance on each dataset is averaged over three independent runs using stratified splits. The error bars represent the average standard deviation across the three runs for the 30 datasets. As shown in the plot, the full setting achieves the highest performance, with an average R-squared value of 0.6288. The removal of regularizations and the replacement of the adaptive margin loss both result in a slight drop in performance. Notably, intra-channel regularization in node aggregation patterns appears to play a more important role in pre-training compared to inter-channel alignment regularization. The largest performance drop is observed in the multi-task learning setting. It is important to note that the only difference between this setting and the multi-channel learning setting is that the latter framework learns the same set of tasks in separate channels. This highlights the effectiveness of multi-channel learning, as it decomposes the pre-trained tasks based on different aspects of chemical knowledge, allowing for the combination of pre-trained knowledge in a task-specific manner. The full performance table is shown in Table~\ref{tab:pretrain_ablation_detail}.

\subsection{Fine-tune strategies}

Additionally, we examine three different strategies for leveraging the learned channel-wise representations during fine-tuning:

\begin{enumerate}
    \item Simple concatenation (Concat) of the channel-wise representations without applying any channel weights. The prediction is made using a task-specific fully connected layer (FC) as the prediction head on top of the concatenated representations.
    \begin{equation}
        \text{prediction} = \text{FC}([h_g^{\text{[MCD]}}, h_g^{\text{[SCD]}}, h_g^{\text{[CP]}}])
    \end{equation}
    
    \item Aggregation of channel-wise representations via a learnable prompt tuning module $\tau_{\theta}(\cdot)$, which corresponds to the primary setup in our experiments. In this case, $\tau_{\theta}$ 
    is a tunable prompt weight (PW) vector consisting of three elements, each representing the logit $l$ for a specific channel. Its initialization is guided by the ROGI value \cite{roughness}. Additionally, a temperature $t$ is used to control the sparsity of the channel selection. A smaller temperature value results in a sharper distribution, concentrating the probability mass on fewer channels. In this ablation, we compare the effect of different temperature values $t \in \{1, 0.7, 0.3\}$. The best performance reported in the main text is achieved using $t = 0.7$. It is important to note that this method is task-specific but sample-agnostic.
    \begin{align}
        \tau_{\theta} = [l_0, l_1, l_2], \:\: \alpha = \text{Softmax} (\tau_{\theta} / t), \:\:\text{prediction} = \text{FC}(\sum_{i} \alpha_i h_g^i)
    \end{align}
    
    \item Aggregation of channel-wise representations using a Mixture-of-Experts \cite{moe} (MoE) alike paradigm. In this approach, $\tau_{\theta}(\cdot)$ is also learned to aggregate the channel-wise representations, but in a different manner. Specifically, for each molecular graph, a gating network (e.g., a fully connected layer) takes the graph representation obtained via mean pooling as input and outputs a set of importance weights for each expert (i.e., channel). Similar to the previous method, the composed representation is a weighted sum of the channel-wise representations. Again, different temperature values $t$ are explored. Notably, this is both a task-specific and sample-specific aggregation method, meaning that channel importance can vary between different molecules. While this approach enhances the model's expressiveness, it also reduces interpretability and increases the risk of overfitting by the additional complexity.
    \begin{align}
        \tau_{\theta} = \text{GATE} (h_g^{mean}), \:\: \alpha = \text{Softmax} (\tau_{\theta} / t), \:\:\text{prediction} = \text{FC}(\sum_{i} \alpha_i h_g^i)
    \end{align}
\end{enumerate}

\begin{figure}[!t]
    \centering
    \includegraphics[width=1\columnwidth, trim={0cm 0cm 0cm 0cm},clip]{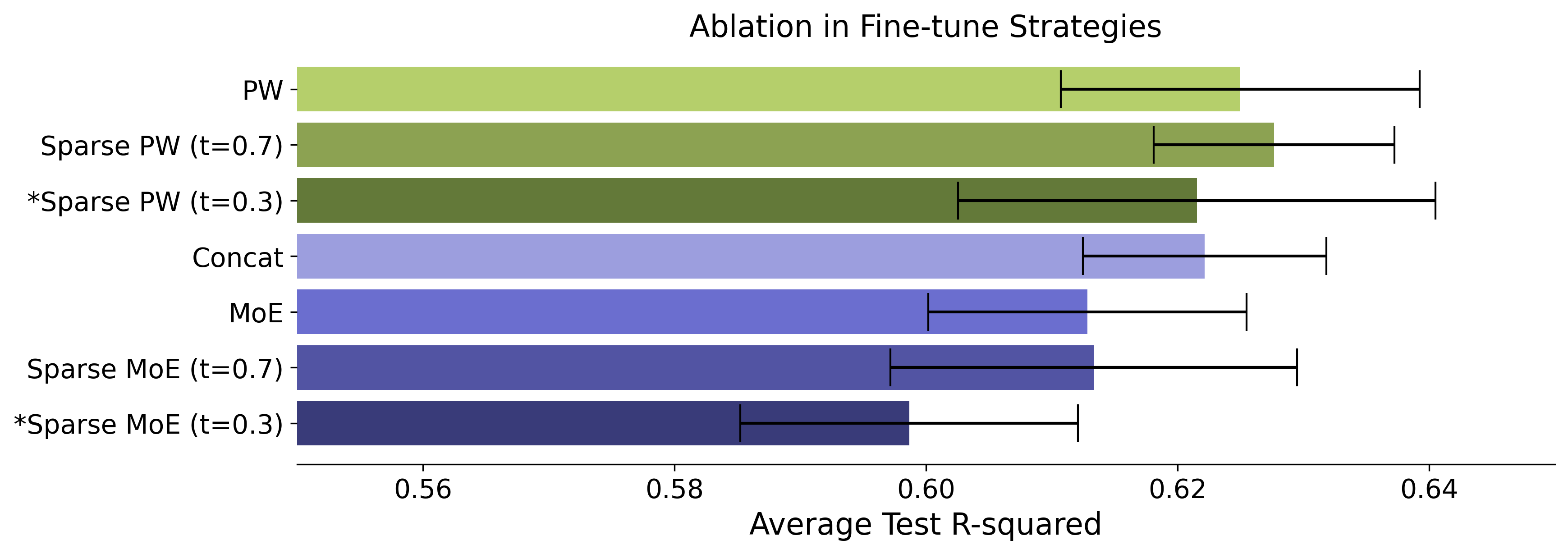}
    \caption{{\textbf{Ablation study in fine-tune strategies.} The average performance on the 30 datasets in MoleculeACE \cite{MoleculeACE} using 7 fine-tune strategies of leveraging the channel-wise representations. The error bars represent the average standard deviation of the standard deviation across the three independent runs for each dataset. See Table~\ref{tab:finetune_ablation_detail} for more details.}} 
    \label{fig:finetune_ablation}
\end{figure}

This experiment is conducted using the best model checkpoint of GIN from the main experiments. The average test R-squared value across the 30 datasets in MoleculeACE was measured, with performance on each dataset averaged over three independent runs using stratified splits. The error bars represent the average standard deviation across the three runs for the 30 datasets. As shown in Figure~\ref{fig:finetune_ablation}, Sparse PW with $t = 0.7$ achieves the best performance. The Concat method, which evenly utilizes all available information, performs slightly worse than PW. This highlights the advantages of selectively utilizing pre-trained information rather than using everything indiscriminately. Surprisingly, the MoE approach performs the worst, despite its ability to select channels via $\tau_{\theta}$. One of the reasons could be its prone to overfitting due to the additional complexity. Additionally, unlike the learnable PW, MoE does not incorporate prior information from the ROGI measure, highlighting the advantages of ROGI-guided initialization. Moreover, it is unsure whether the graph representation derived from mean pooling is suitable for inferring the importance of each channel. It would be interesting to further explore effective aggregation methods for channel-wise representations in future work. Another interesting observation is the tradeoff between information sparsity and completeness. For both PW and MoE, the performance improves from $t=1$ to $t=0.7$ but drops from $t=0.7$ to $t=0.3$. This suggests that some degree of channel selectivity is beneficial. However, as the temperature continues to decrease, the model will tend to rely on a single channel for predictions, ignoring information of other channels. This approach proves to be less effective than leveraging multiple channels. Detailed performance is included in Table~\ref{tab:finetune_ablation_detail}.

\section{Interpretability of channel selectivity}

Since our method can build task-specific, context-dependent representations through channel aggregation, it would be interesting to see how representation from each channel affects the model performance, and whether this aligns with the chemical knowledge required for the downstream property prediction. This analysis mainly tackles the interpretability of the learnable prompt modules during fine-tuning.

\subsection{Case study in channel activation}

Recall that each channel focuses on a specific aspect of the molecule: channel MCD focuses on molecular similarity, channel SCD focuses on scaffold similarity, and channel CP focuses on local patterns like functional group composition. To keep the experiment simple, we focus solely on whether a channel is activated, disregarding the exact weight of each channel. Consequently, we evenly assign the prompt weights across the activated channels. For example, if only MCD and SCD are activated, the prompt weights would be [0.5, 0.5, 0]. As shown in Table~\ref{tab:prompt_weight_ablation}, we choose four binding potency prediction datasets, and run Our$_\text{GPS}$ model on multiple settings with different channel activation. In order to quantify which part of the chemical knowledge is more beneficial for solving the tasks, we decide to analyze the quantitative structure-property relationship (QSPR) between the potency label and the structural features of molecule fingerprint, scaffold fingerprint, and the binary functional group descriptors. We formulate our QSPR metric as the pearson correlation between the fingerprint/descriptor differences (i.e., $1 - \text{Tanimoto similarity}$) and the potency label differences. ECFP4 fingerprint is used. We then normalize the three correlation values. Since this QSPR metric (also true for most QSPR metrics) does not account for distribution shift, it can barely correlate with the model performance comparison under the original stratified sampling data splits \cite{MoleculeACE}. Therefore, the performance in the table corresponds to the average R-squared value across three runs under the random split. By examining the columns where only one channel is activated, the performance shows a high correlation with the QSPR metric. For example, in the CHEMBL236\_Ki dataset, the functional group descriptor has the lowest normalized correlation value of 0.203 with the potency labels. Correspondingly, the R-squared value is lowest when using the CP channel alone. However, when multiple channels are activated, the relationship between channel activation and QSPR metric becomes less correlated. It is not necessarily true that two channels with high corresponding QSPR values will lead to better performance. A typical example is when activating channels MCD and SCD on the CHEMBL237\_Ki dataset. Even though the molecule and scaffold fingerprints show high QSPR values, using both channels worsens the performance. This could be explained by feature redundancy within the two channels that degrades the performance. It also explains why activating all channels simultaneously may not lead to the best performance,  which is consistent with the results in Figure~\ref{fig:finetune_ablation}.

\subsection{Relationship between optimized prompt weights and QSPR metrics}

We further examine the relationship between the optimized prompt weights and the QSPR metrics, which approximate the chemical knowledge required for solving the tasks. The key difference between this analysis and the previous case study is that the optimized prompt weights are fully machine-learned rather than handcrafted. They result from both the ROGI-guided initialization and subsequent optimization. To begin, we retrieve the optimized prompt weights (PW) from the best validation model after fine-tuning on each of the 30 datasets in MoleculeACE using the random split. We then collect the same QSPR measures from above (i.e., the normalized correlation between representation difference and label difference) for all datasets. Finally, we conduct a principal component analysis (PCA) on these two sets of normalized vectors and visualize their relationship in a scatter plot using the computed first components. 

\begin{figure}[!t]
    \centering
    \includegraphics[width=0.7\columnwidth, trim={0cm 0cm 0cm 0cm},clip]{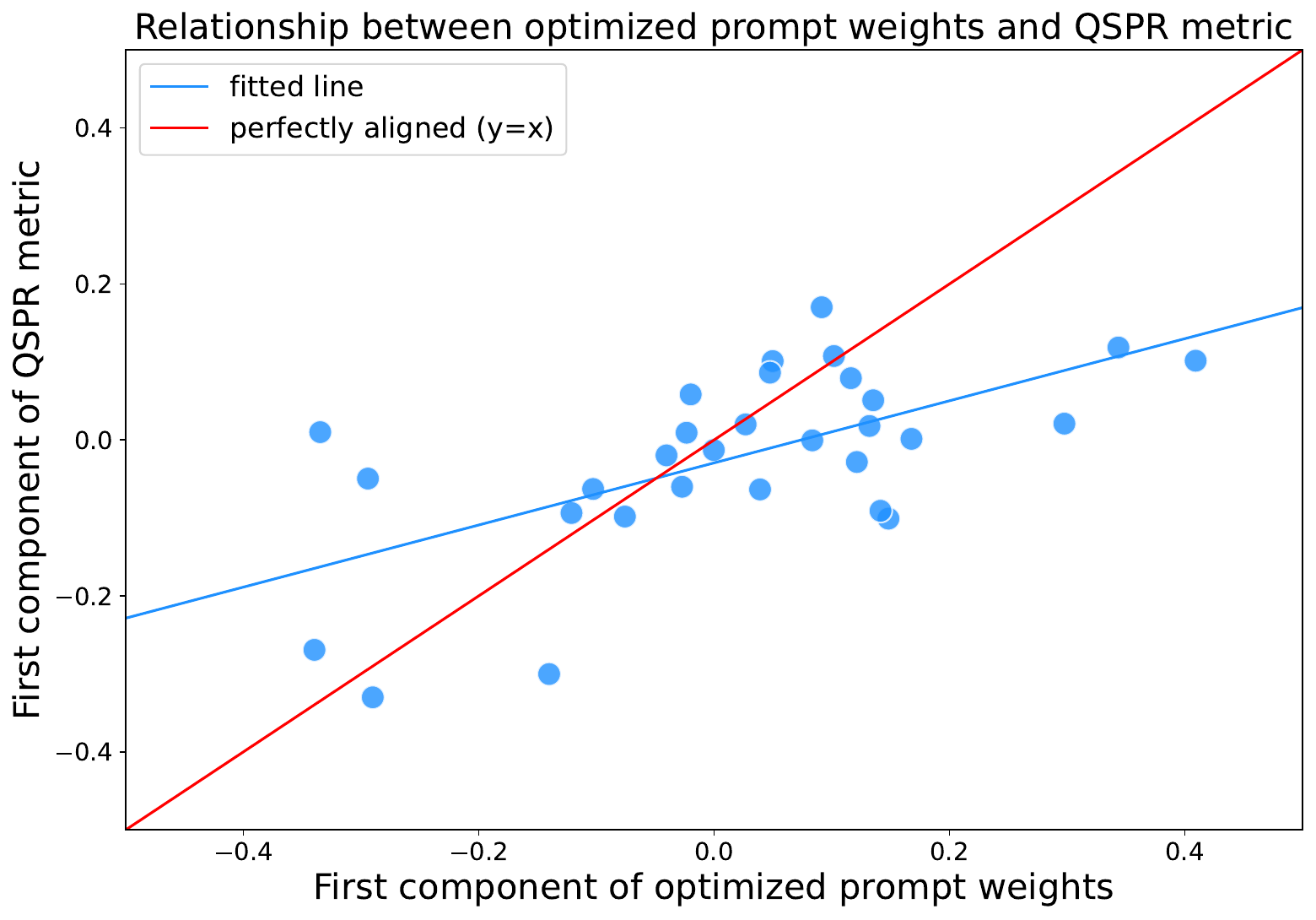}
    \caption{{\textbf{Relationship between optimized prompt weights (PW) and QSPR measure.} The optimized prompt weights is collected from the best validation model during fine-tuning. This experiment is done using random splits on the 30 datasets ($n=30$) in MoleculeACE \cite{MoleculeACE}. QSPR stands for the quantitative structure-property relationship measures we used to approximate the required chemical knowledge for solving the tasks. Principle component analysis (PCA) is performed and the relationship between the first components of PW and QSPR are visualized, as well as the fitted line. }} 
    \label{fig:pw_versus_qspr}
\end{figure}

As shown in Figure~\ref{fig:pw_versus_qspr}, we observe some correlations between the optimized prompt weights (PW) and QSPR measures. However, the alignment is far from perfect. One possible reason could be the feature redundancy hidden within the channel-wise representations, which may influence the model’s decision. Additionally, due to the black-box nature of machine learning models, the problem may be solved in ways that differ from human interpretation. This suggests that hidden information might be processed differently by the model, which requires further investigation. Moreover, it is important to note that large-scale pre-training does not necessarily endow the model with perfect capability in solving the pre-train tasks. Learning multiple tasks simultaneously can still pose challenges, even within the multi-channel learning framework. This indicates that there may be information gaps between the learned channel-wise representations and the actual chemical knowledge they are meant to capture. An intriguing direction for future work would be to analyze how the model’s actual capabilities, as learned during pre-training, influence fine-tuning performance.

\section{Framework extension with other input representations}
As mentioned in the Discussion section of the main text, it would be interesting to apply our framework to input representations other than 2D molecular graphs. However, depending on the downstream application, it is uncertain whether one input representation would outperform the others in terms of model performance. Here, we provide a brief discussion on the possible ways of extending our framework for reference purposes. Two scenarios are discussed: (1) changing the single-modal representation with/without altering the pre-training tasks, and (2) utilizing multi-modal representations along with cross-modal pre-training tasks.

Regarding the single-modal setting (i.e., using only SMILES or 3D graphs), our framework can be easily adapted to both scenarios. The primary adjustment involves changing the encoder accordingly (e.g., a Transformer-based encoder for SMILES and a 3D-GNN for 3D geometry). For the pre-training tasks, special attention must be paid to the definition of subgraph under different modalities. In the case of a SMILES string, its subsequences do not necessarily correspond to valid substructures within the molecule. We would need to extract the subgraph from molecular graph and map it back to the SMILES sequence. For 3D geometry, graphs are often defined differently (e.g., neighbors are defined by geometric distance rather than chemical bonds \cite{multiplex}). Hence, we need to ensure that graph operations are consistent with the given definition. Scaffold-invariant perturbations will not be affected by the change in input representations. Essentially, new molecules are created by the perturbation. On the other hand, there are a lot more self-supervised learning tasks that can be performed with 3D geometry (e.g., 3D coordinates recovery \cite{uni-mol} and bond angle prediction \cite{gem}), which can be incorporated into additional learning channels. However, we believe it is crucial to group these tasks into distinct categories based on commonalities, such as the global-local perspective of the molecule. As suggested in Table~\ref{tab:prompt_weight_ablation}, the fine-tuning performance could be affected by the feature redundancy across channels. 

One potential way to extend the framework with multi-modal representations is to leverage and group the pre-training tasks into distinct channels by information level (e.g., 1D, 2D, or 3D tasks). Additionally, we could incorporate cross-modal self-supervised learning tasks that explicitly guide channel alignment (e.g., contrastive learning between 2D topological graphs and 3D geometry \cite{graphmvp}). One potential advantage is that communication between different encoders may enhance the learning of individual encoders. Moreover, the framework may become more transferable to tasks that require different levels of knowledge (e.g., structure composition versus conformation information).

\clearpage

\clearpage

\clearpage

\begin{figure}[!t]
    \centering
    \includegraphics[width=1\columnwidth, trim={1.2cm 16.7cm 1.2cm 1.3cm},clip]{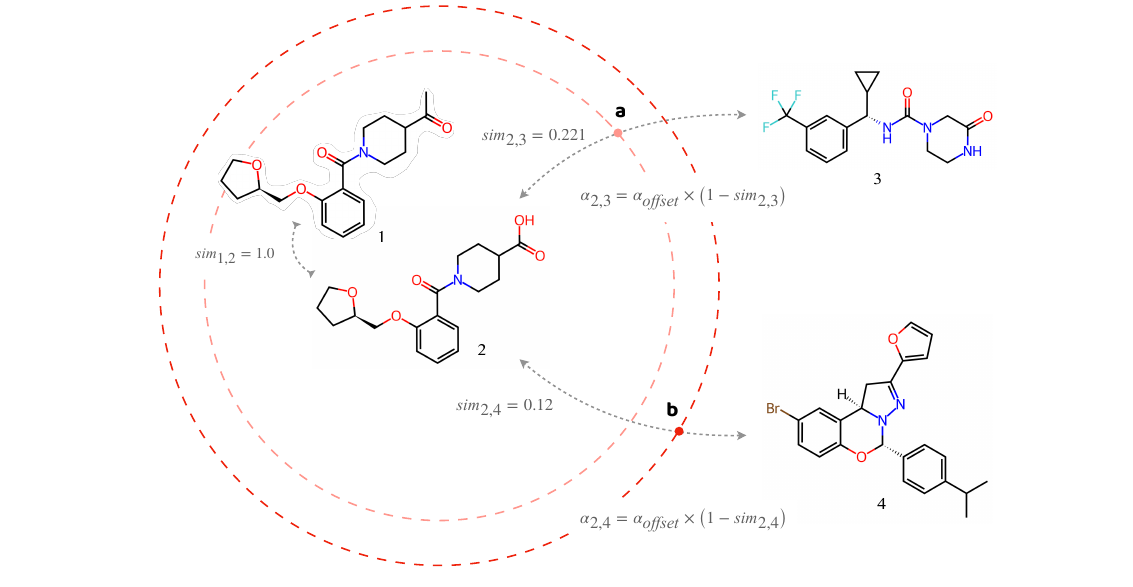}
    \caption{\textbf{Adaptive margin loss.} Triplet examples \textit{$<$1,2,3$>$} and \textit{$<$1,2,4$>$} considered in scaffold distancing adaptive margin loss. $sim_{i, j}$ represents the scaffold fingerprint Tanimoto similarity between compound $i$ and $j$, while $\alpha_{i, j}$ corresponds to the computed margin (indicated by the dashed line). Note that, in this example, (a) the margin $\alpha_{2, 3}$ is smaller than (b) the margin $\alpha_{2, 4}$ because of the similarity difference. Both compound 3 and 4 lie outside of the margin, hence making the triplet loss $\ell_{1, 2, 3} = \ell_{1, 2, 4} = 0$. However, the actual representation distance between \textit{$<$2, 3$>$} and \textit{$<$2, 4$>$} may not necessarily correlate with the computed margins, as shown by the figure. This motivates our quadruplets formulation.}
    \label{fig:margin_loss}
\end{figure}

\clearpage

\clearpage

\clearpage

\clearpage

\clearpage

\clearpage

\clearpage

\clearpage

\begin{figure}[!t]
    \centering
    \includegraphics[width=1\columnwidth, trim={0.3cm 16.7cm 0.3cm 1cm},clip]{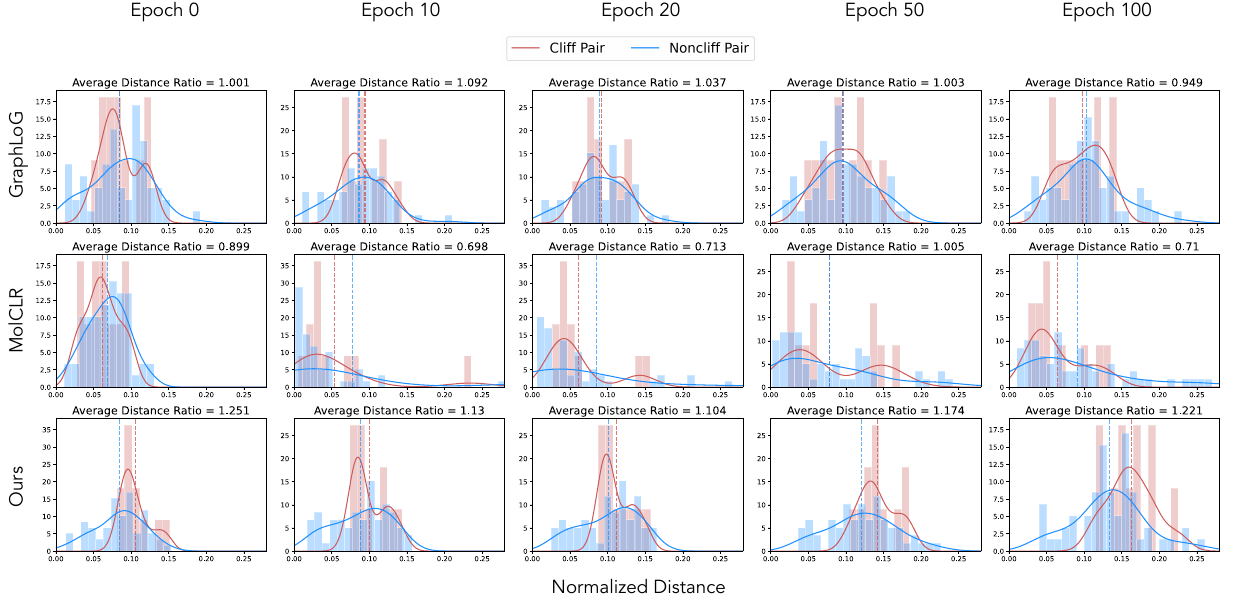}
    \caption{{\textbf{Histogram of distances of cliff and non-cliff pairs.} The shift in cliff distances and non-cliff distances during the representation space probing in Figure 3. For both GraphLoG \cite{graphlog} and our method, non-cliff and cliff matched molecule pairs (MMPs) are pushed further apart along the training process. This is understandable, as the learned representations become more label-oriented rather than structure-oriented during fine-tuning, leading to some loss of structural information. However, it remains important for the model to maintain distance differences for understanding of activity cliffs throughout fine-tuning, which is particularly true for our method. In contrast, MolCLR \cite{molclr} exhibits more oscillatory behavior, with less consistent distance distribution patterns during fine-tuning. This corresponds to the drastic shifts of MolCLR's representation space shown in the main text.}} 
    \label{fig:shift_histogram}
\end{figure}

\clearpage

\begin{figure}[!t]
    \centering
    \includegraphics[width=0.7\columnwidth, trim={0cm 0cm 0cm 0cm},clip]{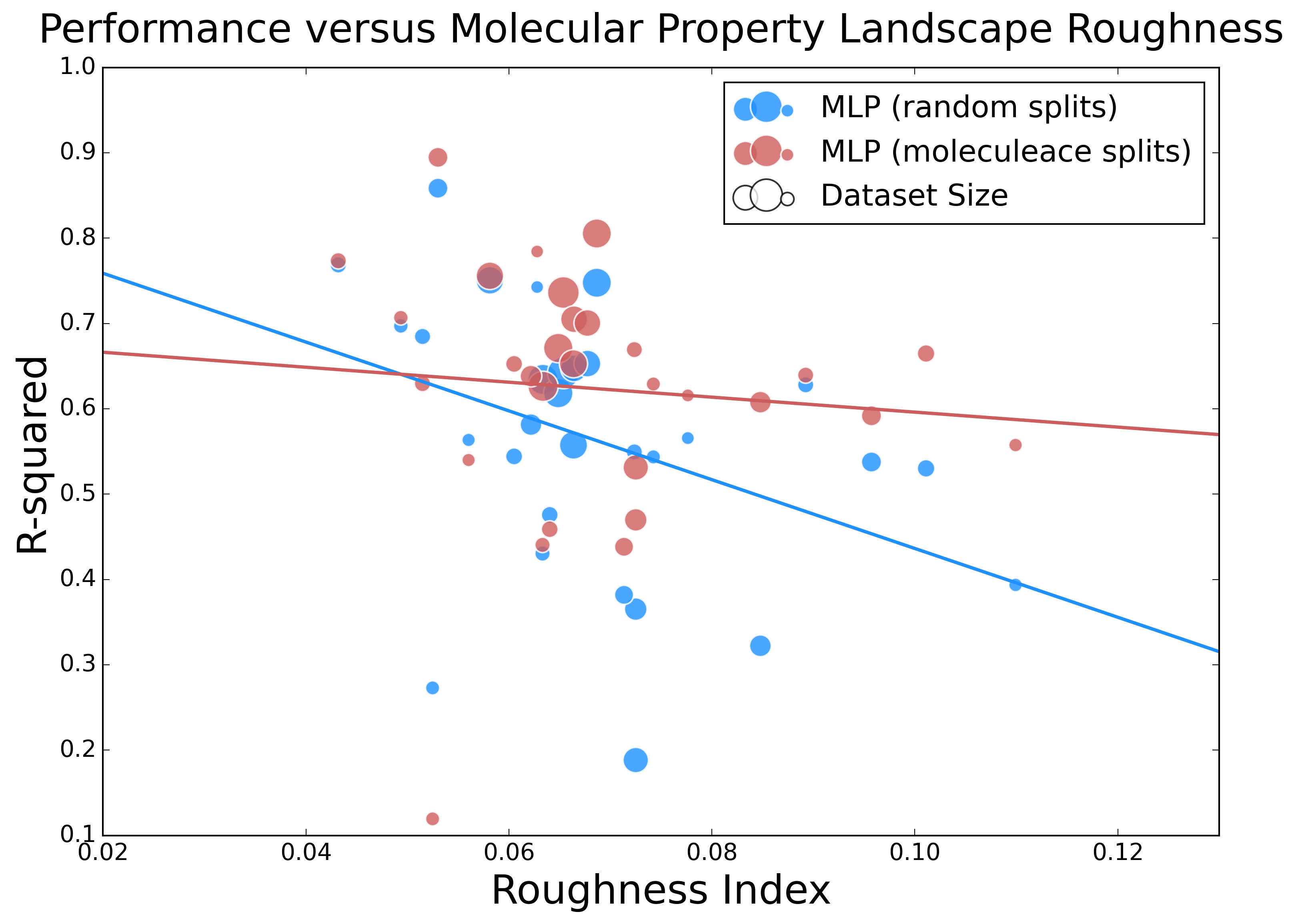}
    \caption{{\textbf{Relationship between ROGI and model performance under different splits.} The test performance of MLP with ECFP4 fingerprint on the 30 datasets ($n=30$) in MoleculeACE \cite{MoleculeACE}, averaged across three independent runs, in relation to the roughness of molecular property landscapes \cite{roughness} under the random splits and the MoleculeACE's stratified splits. The size of the dots represents the dataset size, and the slope of the fitted line indicates the correlation. It demonstrates how the correlation between measures of chemical space discontinuity and the model performance gets affected by the designated data splits. This explains our choice of random splits for all experiments involving QSPR analysis.}} 
    \label{fig:mlp_random_versus_stratified}
\end{figure}

\clearpage

\bgroup
\def\arraystretch{1.2}%
\begin{table}[]
\caption{\textbf{Channel activation ablation.} The performance comparison on CHEMBL236\_Ki, CHEMBL1871\_Ki, CHEMBL228\_Ki, and CHEMBL237\_Ki datasets \cite{MoleculeACE} when enabling different channels. QSPR stands for the quantitative structure-property relationship metric we used. We manually assign the prompt weights to the channels regarding their activation. This experiment is done using random splits.} 
\label{tab:prompt_weight_ablation}
\centering
\begin{tabular}{l|cc|ccccccc}
\Xhline{1.5pt}
\textbf{Dataset} & \multicolumn{1}{c|}{\textbf{QSPR}} & \textbf{Channel} & \multicolumn{7}{c}{\textbf{Activation}}               \\ \hline
CHEMBL236\_Ki    & \multicolumn{1}{c|}{0.423}         & MCD              & $\checkmark$     &       &       & $\checkmark$     &       & $\checkmark$     & $\checkmark$     \\
                 & \multicolumn{1}{c|}{0.374}         & SCD              &       & $\checkmark$     &       & $\checkmark$     & $\checkmark$     &       & $\checkmark$     \\
                 & \multicolumn{1}{c|}{0.203}         & CP               &       &       & $\checkmark$     &       & $\checkmark$     & $\checkmark$     & $\checkmark$     \\ \cline{2-10} 
                 & \multicolumn{2}{c|}{Performance}                      & 0.765 & 0.758 & 0.756 & 0.771 & 0.764 & 0.776 & 0.772 \\ \hline
CHEMBL1871\_Ki   & \multicolumn{1}{c|}{0.391}         & MCD              & $\checkmark$     &       &       & $\checkmark$     &       & $\checkmark$     & $\checkmark$     \\
                 & \multicolumn{1}{c|}{0.272}         & SCD              &       & $\checkmark$     &       & $\checkmark$     & $\checkmark$     &       & $\checkmark$     \\
                 & \multicolumn{1}{c|}{0.338}         & CP               &       &       & $\checkmark$     &       & $\checkmark$     & $\checkmark$     & $\checkmark$     \\ \cline{2-10} 
                 & \multicolumn{2}{c|}{Performance}                      & 0.523 & 0.469 & 0.541 & 0.529 & 0.515 & 0.526 & 0.518 \\ \hline
CHEMBL228\_Ki    & \multicolumn{1}{c|}{0.402}         & MCD              & $\checkmark$     &       &       & $\checkmark$     &       & $\checkmark$     & $\checkmark$     \\
                 & \multicolumn{1}{c|}{0.266}         & SCD              &       & $\checkmark$     &       & $\checkmark$     & $\checkmark$     &       & $\checkmark$     \\
                 & \multicolumn{1}{c|}{0.332}         & CP               &       &       & $\checkmark$     &       & $\checkmark$     & $\checkmark$     & $\checkmark$     \\ \cline{2-10} 
                 & \multicolumn{2}{c|}{Performance}                      & 0.667 & 0.644 & 0.662 & 0.644 & 0.667 & 0.667 & 0.662 \\ \hline
CHEMBL237\_Ki    & \multicolumn{1}{c|}{0.461}         & MCD              & $\checkmark$     &       &       & $\checkmark$     &       & $\checkmark$     & $\checkmark$     \\
                 & \multicolumn{1}{c|}{0.501}         & SCD              &       & $\checkmark$     &       & $\checkmark$     & $\checkmark$     &       & $\checkmark$     \\
                 & \multicolumn{1}{c|}{0.039}         & CP               &       &       & $\checkmark$     &       & $\checkmark$     & $\checkmark$     & $\checkmark$     \\ \cline{2-10} 
                 & \multicolumn{2}{c|}{Performance}                      & 0.769 & 0.772 & 0.762 & 0.749 & 0.765 & 0.756 & 0.757 \\ \Xhline{1.5pt}
\end{tabular}
\end{table}
\egroup

\begin{landscape}
\begin{table}[]
\caption{Fine-tuning performance (R-squared) of 12 methods on 30 binding potency prediction tasks in MoleculeACE \cite{MoleculeACE} {using stratified splits}. C2047$_\text{EC}$ is short for the CHEMBL2047\_EC50 dataset, while GraLoG=GraphLoG, GROV=GROVER, and MoLF=MoLFormer. The performance is averaged across three independent runs, where standard deviation is also shown. Zero standard deviation indicates that the value is below 0.01. Best performance is shown in bold.}
\label{tab:moleculeace}
\begin{adjustbox}{width=\columnwidth, center}
\begin{tabular}{lllllllllllll}
\Xhline{1.5pt} 
\textbf{Dataset} & \textbf{MLP} & \textbf{GIN} \cite{gin} & \textbf{GraLoG} \cite{graphlog} & \textbf{MolCLR} \cite{molclr} & \textbf{GROV} \cite{gnnpretrain_grover} & \textbf{MoLF} \cite{molformer} & \textbf{GEM} \cite{gem} & \textbf{UniMol} \cite{uni-mol} & \textbf{KANO} \cite{kano} & \textbf{KPGT} \cite{kpgt} & \textbf{Ours$_\text{GIN}$} & \textbf{Ours$_\text{GPS}$} \\ \hline
C1862$_\text{Ki}$   & 0.71$\pm$0.01   & 0.71$\pm$0.03   & 0.69$\pm$0.03        & 0.76$\pm$0.03      & 0.75$\pm$0.01      & 0.71$\pm$0.0          & 0.79$\pm$0.01   & 0.81$\pm$0.01       & 0.76$\pm$0.01    & \textbf{0.82$\pm$0.01}    & 0.77$\pm$0.01      & \textbf{0.82$\pm$0.01}      \\
C1871$_\text{Ki}$   & 0.54$\pm$0.02   & 0.42$\pm$0.05   & 0.33$\pm$0.03        & 0.41$\pm$0.09      & 0.42$\pm$0.06      & 0.47$\pm$0.0          & 0.46$\pm$0.04   & 0.49$\pm$0.06        & 0.55$\pm$0.02    & \textbf{0.63$\pm$0.01}    & 0.54$\pm$0.01      & 0.59$\pm$0.01      \\
C2034$_\text{Ki}$   & \textbf{0.63$\pm$0.01}   & 0.3$\pm$0.06    & 0.56$\pm$0.03        & 0.4$\pm$0.1        & 0.52$\pm$0.02      & 0.6$\pm$0.02          & 0.44$\pm$0.04   & 0.34$\pm$0.04       & 0.5$\pm$0.04     & 0.55$\pm$0.03    & 0.52$\pm$0.02       & 0.55$\pm$0.04      \\
C2047$_\text{EC}$ & \textbf{0.62$\pm$0.01}   & 0.41$\pm$0.08   & 0.52$\pm$0.02        & 0.44$\pm$0.09      & 0.59$\pm$0.02      & 0.6$\pm$0.02          & 0.38$\pm$0.02   & 0.34$\pm$0.08       & 0.25$\pm$0.11    & 0.25$\pm$0.02    & 0.32$\pm$0.05      & 0.31$\pm$0.02      \\
C204$_\text{Ki}$    & 0.76$\pm$0.01   & 0.76$\pm$0.02   & 0.71$\pm$0.02        & 0.72$\pm$0.02      & 0.66$\pm$0.01      & 0.73$\pm$0.0          & 0.78$\pm$0.01   & 0.81$\pm$0.01       & 0.77$\pm$0.01    & 0.82$\pm$0.01    & 0.79$\pm$0.01      & \textbf{0.83$\pm$0.0}      \\
C2147$_\text{Ki}$   & 0.89$\pm$0.0    & 0.88$\pm$0.01   & 0.82$\pm$0.01        & 0.87$\pm$0.02      & 0.85$\pm$0.01      & 0.86$\pm$0.01         & 0.85$\pm$0.01   & 0.87$\pm$0.01       & 0.86$\pm$0.01    & \textbf{0.9$\pm$0.0}      & 0.87$\pm$0.0      & 0.89$\pm$0.0       \\
C214$_\text{Ki}$    & 0.63$\pm$0.0    & 0.63$\pm$0.01   & 0.58$\pm$0.02        & 0.6$\pm$0.02       & 0.45$\pm$0.09      & 0.59$\pm$0.02         & 0.6$\pm$0.03    & 0.66$\pm$0.01       & 0.61$\pm$0.03    & \textbf{0.68$\pm$0.01}    & 0.64$\pm$0.01      & 0.67$\pm$0.01      \\
C218$_\text{EC}$  & 0.46$\pm$0.03   & 0.35$\pm$0.06   & 0.34$\pm$0.03        & 0.37$\pm$0.04      & 0.48$\pm$0.07      & 0.23$\pm$0.05         & 0.52$\pm$0.04   & 0.56$\pm$0.07       & 0.59$\pm$0.02    & \textbf{0.65$\pm$0.01}    & 0.58$\pm$0.04      & 0.65$\pm$0.02      \\
C219$_\text{Ki}$    & 0.47$\pm$0.02   & 0.44$\pm$0.04   & 0.4$\pm$0.04         & 0.37$\pm$0.04      & 0.32$\pm$0.05      & 0.46$\pm$0.02         & 0.25$\pm$0.03   & 0.40$\pm$0.02       & 0.32$\pm$0.01    & \textbf{0.48$\pm$0.01}    & 0.31$\pm$0.04      & 0.35$\pm$0.03      \\
C228$_\text{Ki}$    & 0.64$\pm$0.01   & 0.63$\pm$0.02   & 0.52$\pm$0.03        & 0.58$\pm$0.03      & 0.46$\pm$0.11      & 0.64$\pm$0.01         & 0.64$\pm$0.02   & 0.68$\pm$0.02       & 0.65$\pm$0.02    & 0.68$\pm$0.01    & 0.68$\pm$0.01       & \textbf{0.7$\pm$0.01}       \\
C231$_\text{Ki}$    & 0.63$\pm$0.01   & 0.58$\pm$0.02   & 0.54$\pm$0.03        & 0.57$\pm$0.03      & 0.64$\pm$0.02      & 0.55$\pm$0.0          & 0.8$\pm$0.02    & \textbf{0.81$\pm$0.01}       & 0.77$\pm$0.03    & 0.79$\pm$0.01    & 0.74$\pm$0.0       & 0.79$\pm$0.02      \\
C233$_\text{Ki}$    & 0.67$\pm$0.0    & 0.63$\pm$0.01   & 0.59$\pm$0.02        & 0.64$\pm$0.02      & 0.5$\pm$0.09       & 0.7$\pm$0.0           & 0.67$\pm$0.02   & 0.72$\pm$0.01       & 0.69$\pm$0.0     & 0.71$\pm$0.01    & 0.71$\pm$0.01       & \textbf{0.72$\pm$0.0}      \\
C234$_\text{Ki}$    & \textbf{0.74$\pm$0.0}    & 0.66$\pm$0.02   & 0.6$\pm$0.01         & 0.62$\pm$0.04      & 0.49$\pm$0.01      & 0.71$\pm$0.01         & 0.63$\pm$0.01   & 0.64$\pm$0.02       & 0.67$\pm$0.02    & 0.7$\pm$0.01     & 0.65$\pm$0.0       & 0.69$\pm$0.01       \\
C235$_\text{EC}$  & 0.53$\pm$0.01   & 0.52$\pm$0.02   & 0.51$\pm$0.04        & 0.48$\pm$0.03      & 0.4$\pm$0.1        & 0.49$\pm$0.02         & 0.66$\pm$0.02   & 0.69$\pm$0.02       & 0.69$\pm$0.02    & 0.72$\pm$0.01    & 0.7$\pm$0.02      & 0.\textbf{74$\pm$0.01}      \\
C236$_\text{Ki}$    & 0.7$\pm$0.01    & 0.7$\pm$0.02    & 0.65$\pm$0.02        & 0.68$\pm$0.02      & 0.52$\pm$0.09      & 0.65$\pm$0.02         & 0.64$\pm$0.05   & 0.74$\pm$0.0       & 0.67$\pm$0.01    & \textbf{0.77$\pm$0.01}    & 0.74$\pm$0.01      & 0.74$\pm$0.01      \\
C237$_\text{EC}$  & 0.64$\pm$0.01   & 0.6$\pm$0.05    & 0.51$\pm$0.03        & 0.53$\pm$0.07      & 0.52$\pm$0.11      & 0.6$\pm$0.0           & 0.64$\pm$0.02   & 0.62$\pm$0.01       & 0.66$\pm$0.02    & 0.71$\pm$0.01    & 0.66$\pm$0.01      & \textbf{0.72$\pm$0.03}      \\
C237$_\text{Ki}$    & 0.7$\pm$0.0     & 0.66$\pm$0.01   & 0.67$\pm$0.02        & 0.67$\pm$0.02      & 0.6$\pm$0.06       & 0.67$\pm$0.03         & 0.72$\pm$0.01   & 0.71$\pm$0.0       & 0.68$\pm$0.0     & \textbf{0.75$\pm$0.01}    & 0.71$\pm$0.0      & 0.7$\pm$0.01       \\
C238$_\text{Ki}$    & 0.65$\pm$0.01   & 0.62$\pm$0.04   & 0.56$\pm$0.02        & 0.63$\pm$0.06      & 0.62$\pm$0.01      & 0.64$\pm$0.01         & 0.63$\pm$0.03   & 0.71$\pm$0.0       & 0.71$\pm$0.01    & \textbf{0.73$\pm$0.0}     & 0.68$\pm$0.02       & 0.7$\pm$0.03       \\
C239$_\text{EC}$  & \textbf{0.61$\pm$0.0}    & 0.47$\pm$0.06   & 0.41$\pm$0.04        & 0.47$\pm$0.07      & 0.52$\pm$0.02      & 0.58$\pm$0.02         & 0.5$\pm$0.01    & 0.48$\pm$0.0        & 0.44$\pm$0.03    & 0.54$\pm$0.02    & 0.58$\pm$0.02       & 0.58$\pm$0.02      \\
C244$_\text{Ki}$    & 0.81$\pm$0.0    & 0.8$\pm$0.01    & 0.78$\pm$0.01        & 0.8$\pm$0.02       & 0.63$\pm$0.06      & 0.79$\pm$0.0          & 0.76$\pm$0.01   & 0.8$\pm$0.0        & 0.78$\pm$0.01    & \textbf{0.82$\pm$0.0}     & 0.81$\pm$0.0       & 0.81$\pm$0.0       \\
C262$_\text{Ki}$    & 0.44$\pm$0.04   & 0.2$\pm$0.1     & 0.34$\pm$0.04        & 0.45$\pm$0.02      & 0.4$\pm$0.07       & 0.58$\pm$0.01         & 0.51$\pm$0.03   & 0.63$\pm$0.0       & 0.6$\pm$0.02     & \textbf{0.65$\pm$0.01}    & 0.6$\pm$0.03      & 0.64$\pm$0.03      \\
C264$_\text{Ki}$    & 0.65$\pm$0.01   & 0.65$\pm$0.03   & 0.6$\pm$0.02         & 0.62$\pm$0.03      & 0.41$\pm$0.04      & 0.67$\pm$0.01         & 0.62$\pm$0.02   & 0.72$\pm$0.0       & 0.66$\pm$0.01    & \textbf{0.73$\pm$0.01}    & 0.7$\pm$0.0      & \textbf{0.73$\pm$0.01}      \\
C2835$_\text{Ki}$   & 0.78$\pm$0.01   & 0.78$\pm$0.03   & 0.69$\pm$0.04        & 0.75$\pm$0.04      & 0.7$\pm$0.03       & 0.72$\pm$0.01         & 0.79$\pm$0.01   & 0.79$\pm$0.0       & 0.78$\pm$0.02    & 0.8$\pm$0.02     & 0.76$\pm$0.01       & \textbf{0.82$\pm$0.02}      \\
C287$_\text{Ki}$    & 0.44$\pm$0.01   & 0.46$\pm$0.03   & 0.21$\pm$0.04        & 0.42$\pm$0.05      & 0.39$\pm$0.02      & 0.46$\pm$0.04         & 0.45$\pm$0.02   & 0.49$\pm$0.0       & 0.34$\pm$0.05    & \textbf{0.57$\pm$0.01}    & 0.5$\pm$0.01      & 0.52$\pm$0.0      \\
C2971$_\text{Ki}$   & 0.77$\pm$0.02   & 0.72$\pm$0.01   & 0.68$\pm$0.02        & 0.76$\pm$0.06      & 0.78$\pm$0.01      & 0.78$\pm$0.01         & 0.83$\pm$0.03   & 0.81$\pm$0.0        & 0.82$\pm$0.01    & 0.84$\pm$0.02    & 0.85$\pm$0.02      & \textbf{0.85$\pm$0.01}      \\
C3979$_\text{EC}$ & \textbf{0.66$\pm$0.01}   & 0.36$\pm$0.04   & 0.29$\pm$0.03        & 0.35$\pm$0.07      & 0.34$\pm$0.05      & 0.62$\pm$0.05         & 0.47$\pm$0.02   & 0.26$\pm$0.01       & 0.5$\pm$0.01     & 0.5$\pm$0.04     & 0.48$\pm$0.04      & 0.55$\pm$0.05      \\
C4005$_\text{Ki}$   & \textbf{0.67$\pm$0.01}   & 0.64$\pm$0.03   & 0.57$\pm$0.01        & 0.44$\pm$0.06      & 0.58$\pm$0.07      & 0.53$\pm$0.04         & 0.57$\pm$0.02   & 0.49$\pm$0.01       & 0.58$\pm$0.04    & 0.63$\pm$0.01    & 0.55$\pm$0.02      & 0.59$\pm$0.02      \\
C4203$_\text{Ki}$   & 0.12$\pm$0.01   & 0.09$\pm$0.12   & 0.05$\pm$0.02        & 0.09$\pm$0.05      & 0.16$\pm$0.03      & 0.14$\pm$0.0          & 0.16$\pm$0.04   & 0.11$\pm$0.0       & 0.15$\pm$0.05    & 0.33$\pm$0.06    & 0.38$\pm$0.04       & \textbf{0.39$\pm$0.03}       \\
C4616$_\text{EC}$ & \textbf{0.56$\pm$0.01}   & 0.42$\pm$0.06   & 0.49$\pm$0.03        & 0.48$\pm$0.06      & 0.44$\pm$0.1       & 0.55$\pm$0.02         & 0.43$\pm$0.04   & 0.47$\pm$0.0       & 0.53$\pm$0.05    & 0.52$\pm$0.04    & 0.49$\pm$0.02       & 0.48$\pm$0.05      \\
C4792$_\text{Ki}$   & 0.59$\pm$0.01   & \textbf{0.65$\pm$0.02}   & 0.55$\pm$0.03        & 0.62$\pm$0.04      & 0.55$\pm$0.09      & 0.64$\pm$0.0          & 0.51$\pm$0.03   & 0.49$\pm$0.01       & 0.49$\pm$0.02    & 0.55$\pm$0.01    & 0.52$\pm$0.0      & 0.54$\pm$0.03       \\ \hline
Average          & 0.6235       & 0.558        & 0.5253            & 0.553           & 0.523           & 0.5982             & 0.589        & 0.6047          & 0.6025        & \textbf{0.6614}        & 0.6279          & 0.656         \\ \Xhline{1.5pt} 
\end{tabular}
\end{adjustbox}
\end{table}
\end{landscape}

\begin{table}[]
\caption{{Fine-tuning performance (R-squared) of MLP, KPGT \cite{kpgt}, and Our$_\text{GPS}$ on 30 binding potency prediction tasks in MoleculeACE \cite{MoleculeACE} using random splits. The performance is averaged across three independent runs, where standard deviation is also shown. Zero standard deviation indicates that the value is below 0.01. Best performance is shown in bold.}}
\label{tab:moleculeace_random_splits}
\centering
\begin{tabular}{llll}
\Xhline{1.5pt}
\textbf{Dataset} & \textbf{MLP} & \textbf{KPGT \cite{kpgt}} & \textbf{Ours$_\text{GPS}$} \\ \hline
C1862$_\text{Ki}$   & 0.7$\pm$0.05    & 0.74$\pm$0.03    & \textbf{0.81$\pm$0.0}    \\
C1871$_\text{Ki}$   & 0.56$\pm$0.06   & \textbf{0.62$\pm$0.03}    & 0.54$\pm$0.03   \\
C2034$_\text{Ki}$   & 0.54$\pm$0.05   & 0.54$\pm$0.09    & \textbf{0.65$\pm$0.02}   \\
C2047$_\text{EC}$ & \textbf{0.57$\pm$0.06}   & 0.34$\pm$0.21    & 0.52$\pm$0.04   \\
C204$_\text{Ki}$    & 0.75$\pm$0.02   & 0.69$\pm$0.01    & \textbf{0.79$\pm$0.01}   \\
C2147$_\text{Ki}$   & 0.86$\pm$0.01   & 0.85$\pm$0.02    & \textbf{0.91$\pm$0.01}   \\
C214$_\text{Ki}$    & 0.63$\pm$0.03   & 0.62$\pm$0.02    & \textbf{0.7$\pm$0.01}    \\
C218$_\text{EC}$  & \textbf{0.48$\pm$0.05}   & 0.46$\pm$0.01    & 0.43$\pm$0.03   \\
C219$_\text{Ki}$    & 0.37$\pm$0.05   & 0.47$\pm$0.02    & \textbf{0.54$\pm$0.02}   \\
C228$_\text{Ki}$    & 0.58$\pm$0.04   & 0.57$\pm$0.03    & \textbf{0.63$\pm$0.02}   \\
C231$_\text{Ki}$    & 0.68$\pm$0.02   & 0.62$\pm$0.1     & \textbf{0.73$\pm$0.02}   \\
C233$_\text{Ki}$    & 0.62$\pm$0.02   & 0.6$\pm$0.02     & \textbf{0.72$\pm$0.01}   \\
C234$_\text{Ki}$    & 0.64$\pm$0.03   & 0.63$\pm$0.02    & \textbf{0.77$\pm$0.01}   \\
C235$_\text{EC}$  & 0.19$\pm$0.21   & 0.53$\pm$0.04    & \textbf{0.68$\pm$0.01}   \\
C236$_\text{Ki}$    & 0.65$\pm$0.08   & 0.69$\pm$0.02    & \textbf{0.75$\pm$0.01}   \\
C237$_\text{EC}$  & 0.63$\pm$0.04   & 0.41$\pm$0.09    & \textbf{0.68$\pm$0.02}   \\
C237$_\text{Ki}$    & 0.65$\pm$0.02   & 0.65$\pm$0.01    & \textbf{0.76$\pm$0.01}   \\
C238$_\text{Ki}$    & 0.54$\pm$0.07   & \textbf{0.6$\pm$0.02}     & 0.53$\pm$0.03   \\
C239$_\text{EC}$  & 0.32$\pm$0.12   & \textbf{0.57$\pm$0.01}    & 0.56$\pm$0.01   \\
C244$_\text{Ki}$    & 0.75$\pm$0.02   & 0.71$\pm$0.0     & \textbf{0.81$\pm$0.01}   \\
C262$_\text{Ki}$    & 0.43$\pm$0.08   & 0.44$\pm$0.09    & \textbf{0.55$\pm$0.01}   \\
C264$_\text{Ki}$    & 0.56$\pm$0.04   & 0.67$\pm$0.01    & \textbf{0.69$\pm$0.01}   \\
C2835$_\text{Ki}$   & 0.74$\pm$0.04   & \textbf{0.79$\pm$0.04}    & 0.78$\pm$0.01   \\
C287$_\text{Ki}$    & 0.38$\pm$0.11   & 0.53$\pm$0.09    & \textbf{0.74$\pm$0.01}   \\
C2971$_\text{Ki}$   & 0.77$\pm$0.03   & 0.74$\pm$0.03    & \textbf{0.85$\pm$0.02}   \\
C3979$_\text{EC}$ & 0.53$\pm$0.05   & 0.4$\pm$0.09     & \textbf{0.61$\pm$0.03}   \\
C4005$_\text{Ki}$   & 0.55$\pm$0.08   & 0.56$\pm$0.02    & \textbf{0.59$\pm$0.01}   \\
C4203$_\text{Ki}$   & 0.27$\pm$0.07   & \textbf{0.3$\pm$0.15}     & 0.24$\pm$0.05   \\
C4616$_\text{EC}$ & 0.39$\pm$0.09   & 0.31$\pm$0.07    & \textbf{0.48$\pm$0.04}   \\
C4792$_\text{Ki}$   & 0.54$\pm$0.11   & 0.41$\pm$0.16    & \textbf{0.75$\pm$0.0}    \\ \hline
Average          & 0.5625       & 0.5685        & \textbf{0.659}        \\ \Xhline{1.5pt}
\end{tabular}
\end{table}

\begin{landscape}
\begin{table}[]
\caption{{Ablation performance (R-squared) from 6 pre-train settings on 30 binding potency prediction tasks in MoleculeACE \cite{MoleculeACE} using stratified splits. This experiment is performed using GIN \cite{gin} as model backbone. The performance is averaged across three independent runs, where standard deviation is also shown. Zero standard deviation indicates that the value is below 0.01. Best performance is shown in bold.}}
\label{tab:pretrain_ablation_detail}
\begin{adjustbox}{width=0.8\columnwidth, center}
\begin{tabular}{llllllll}
\Xhline{1.5pt}
\textbf{Dataset} & \textbf{Full} & \textbf{\begin{tabular}[c]{@{}l@{}}No \\ regularizations\end{tabular}} & \textbf{\begin{tabular}[c]{@{}l@{}}No intra-channel \\ regularization\end{tabular}} & \textbf{\begin{tabular}[c]{@{}l@{}}No inter-channel \\ regularization\end{tabular}} & \textbf{\begin{tabular}[c]{@{}l@{}}Replace with \\ margin loss\end{tabular}} & \textbf{\begin{tabular}[c]{@{}l@{}}Replace with \\ multi-task learning\end{tabular}} \\ \hline
C1862$_\text{Ki}$   & \textbf{0.79$\pm$0.01}    & 0.74$\pm$0.02                  & 0.77$\pm$0.02                               & 0.74$\pm$0.02                               & 0.77$\pm$0.02                        & 0.75$\pm$0.01                                \\
C1871$_\text{Ki}$   & 0.51$\pm$0.03    & 0.47$\pm$0.01                  & 0.54$\pm$0.04                               & \textbf{0.57$\pm$0.01}                               & 0.43$\pm$0.02                        & 0.46$\pm$0.02                                \\
C2034$_\text{Ki}$   & \textbf{0.62$\pm$0.02}    & 0.57$\pm$0.02                  & 0.6$\pm$0.03                                & 0.52$\pm$0.02                               & 0.56$\pm$0.05                        & 0.58$\pm$0.01                                \\
C2047$_\text{EC}$ & 0.24$\pm$0.05    & 0.35$\pm$0.01                  & 0.37$\pm$0.06                               & \textbf{0.38$\pm$0.09}                               & 0.24$\pm$0.04                        & 0.2$\pm$0.03                                 \\
C204$_\text{Ki}$    & \textbf{0.8$\pm$0.01}     & 0.79$\pm$0.01                  & \textbf{0.8$\pm$0.0}                      & \textbf{0.8$\pm$0.01}           & \textbf{0.8$\pm$0.0}                          & 0.78$\pm$0.01                                \\
C2147$_\text{Ki}$   & 0.86$\pm$0.0     & \textbf{0.88$\pm$0.0}                   & \textbf{0.88$\pm$0.01}                               & 0.85$\pm$0.0                                & 0.87$\pm$0.0                         & 0.87$\pm$0.0                                 \\
C214$_\text{Ki}$    & \textbf{0.65$\pm$0.01}    & 0.62$\pm$0.01                  & 0.63$\pm$0.01                               & \textbf{0.65$\pm$0.02}                               & 0.62$\pm$0.01                        & 0.64$\pm$0.0                                 \\
C218$_\text{EC}$  & 0.48$\pm$0.01    & 0.52$\pm$0.01                  & 0.5$\pm$0.03                                & 0.5$\pm$0.02                                & \textbf{0.59$\pm$0.02}                        & 0.5$\pm$0.01                                 \\
C219$_\text{Ki}$    & 0.38$\pm$0.01    & 0.32$\pm$0.01                  & 0.27$\pm$0.03                               & 0.32$\pm$0.02                               & \textbf{0.39$\pm$0.02}                        & 0.28$\pm$0.01                                \\
C228$_\text{Ki}$    & 0.65$\pm$0.01    & 0.64$\pm$0.01                  & 0.61$\pm$0.0                                & 0.6$\pm$0.01                                & \textbf{0.66$\pm$0.0}                         & 0.63$\pm$0.0                                 \\
C231$_\text{Ki}$    & 0.75$\pm$0.02    & 0.75$\pm$0.02                  & 0.75$\pm$0.03                               & \textbf{0.76$\pm$0.02}                               & 0.75$\pm$0.01                        & 0.75$\pm$0.03                                \\
C233$_\text{Ki}$    & \textbf{0.71$\pm$0.01}    & 0.7$\pm$0.01                   & 0.69$\pm$0.01                               & 0.69$\pm$0.01                               & 0.7$\pm$0.01                         & \textbf{0.71$\pm$0.01}                                \\
C234$_\text{Ki}$    & 0.67$\pm$0.0     & \textbf{0.68$\pm$0.01}                  & 0.64$\pm$0.01                               & 0.65$\pm$0.02                               & \textbf{0.68$\pm$0.0}                         & 0.66$\pm$0.01                                \\
C235$_\text{EC}$  & \textbf{0.71$\pm$0.01}    & 0.66$\pm$0.02                  & 0.67$\pm$0.02                               & 0.65$\pm$0.01                               & 0.68$\pm$0.03                        & 0.66$\pm$0.03                                \\
C236$_\text{Ki}$    & 0.72$\pm$0.01    & \textbf{0.74$\pm$0.0}                   & 0.72$\pm$0.0                                & 0.71$\pm$0.01                               & 0.71$\pm$0.01                        & 0.73$\pm$0.01                                \\
C237$_\text{EC}$  & \textbf{0.68$\pm$0.02}    & 0.66$\pm$0.02                  & 0.64$\pm$0.02                               & 0.66$\pm$0.03                               & 0.64$\pm$0.02                        & 0.64$\pm$0.02                                \\
C237$_\text{Ki}$    & 0.71$\pm$0.0     & 0.71$\pm$0.01                  & 0.69$\pm$0.01                               & 0.7$\pm$0.01                                & \textbf{0.72$\pm$0.01}                        & \textbf{0.72$\pm$0.01}                                \\
C238$_\text{Ki}$    & 0.67$\pm$0.01    & 0.68$\pm$0.01                  & \textbf{0.7$\pm$0.01}                                & 0.67$\pm$0.02                               & 0.63$\pm$0.01                        & 0.63$\pm$0.04                                \\
C239$_\text{EC}$  & 0.48$\pm$0.01    & \textbf{0.5$\pm$0.01}                   & 0.48$\pm$0.03                               & 0.45$\pm$0.02                               & 0.45$\pm$0.01                        & 0.48$\pm$0.02                                \\
C244$_\text{Ki}$    & 0.81$\pm$0.0     & 0.76$\pm$0.01                  & 0.78$\pm$0.01                               & 0.79$\pm$0.01                               & \textbf{0.82$\pm$0.0}                         & 0.81$\pm$0.01                                \\
C262$_\text{Ki}$    & 0.63$\pm$0.02    & \textbf{0.65$\pm$0.02}                  & 0.57$\pm$0.02                               & \textbf{0.65$\pm$0.0}                                & 0.59$\pm$0.02                        & 0.52$\pm$0.02                                \\
C264$_\text{Ki}$    & 0.69$\pm$0.01    & 0.69$\pm$0.02                  & 0.68$\pm$0.01                               & \textbf{0.7$\pm$0.01}                                & 0.69$\pm$0.02                        & 0.66$\pm$0.01                                \\
C2835$_\text{Ki}$   & 0.8$\pm$0.01     & \textbf{0.82$\pm$0.0}                   & 0.78$\pm$0.01                               & 0.77$\pm$0.01                               & \textbf{0.82$\pm$0.01}                        & 0.78$\pm$0.02                                \\
C287$_\text{Ki}$    & 0.53$\pm$0.01    & 0.55$\pm$0.0                   & 0.5$\pm$0.02                                & 0.51$\pm$0.02                               & 0.53$\pm$0.01                        & \textbf{0.56$\pm$0.01}                                \\
C2971$_\text{Ki}$   & 0.83$\pm$0.01    & 0.83$\pm$0.0                   & 0.8$\pm$0.01                                & \textbf{0.85$\pm$0.01}                               & \textbf{0.85$\pm$0.01}                        & 0.81$\pm$0.0                                 \\
C3979$_\text{EC}$ & 0.52$\pm$0.02    & 0.45$\pm$0.02                  & 0.5$\pm$0.01                                & \textbf{0.55$\pm$0.02}                               & 0.52$\pm$0.02                        & 0.52$\pm$0.03                                \\
C4005$_\text{Ki}$   & \textbf{0.59$\pm$0.01}    & 0.58$\pm$0.01                  & 0.56$\pm$0.03                               & 0.56$\pm$0.01                               & \textbf{0.59$\pm$0.0}                         & 0.55$\pm$0.02                                \\
C4203$_\text{Ki}$   & \textbf{0.32$\pm$0.02}    & 0.26$\pm$0.04                  & 0.28$\pm$0.05                               & 0.28$\pm$0.01                               & 0.24$\pm$0.03                        & 0.21$\pm$0.03                                \\
C4616$_\text{EC}$ & \textbf{0.49$\pm$0.03}    & 0.38$\pm$0.02                  & 0.44$\pm$0.01                               & 0.4$\pm$0.01                                & 0.44$\pm$0.02                        & \textbf{0.49$\pm$0.04}                                \\
C4792$_\text{Ki}$   & 0.55$\pm$0.01    & 0.53$\pm$0.01                  & 0.48$\pm$0.04                               & 0.55$\pm$0.03                               & 0.55$\pm$0.01                        & \textbf{0.56$\pm$0.01}                                \\ \hline
Average          & \textbf{0.6288}        & 0.6169                      & 0.6112                                   & 0.6162                                   & 0.6173                            & 0.6045                                    \\ \Xhline{1.5pt}
\end{tabular}
\end{adjustbox}
\end{table}
\end{landscape}

\begin{landscape}
\begin{table}[]
\caption{{Ablation performance (R-squared) of 7 fine-tune strategies on 30 binding potency prediction tasks in MoleculeACE \cite{MoleculeACE} using stratified splits. This experiment is performed using GIN \cite{gin} as model backbone. The performance is averaged across three independent runs, where standard deviation is also shown. Zero standard deviation indicates that the value is below 0.01. Best performance is shown in bold.}}
\label{tab:finetune_ablation_detail}
\begin{adjustbox}{width=0.8\columnwidth, center}
\begin{tabular}{llllllll}
\Xhline{1.5pt}
\multirow{2}{*}{\textbf{Dataset}} & \textbf{PW}  & \textbf{Sparse PW} & \textbf{Sparse PW} & \textbf{Concat} & \textbf{MoE} & \textbf{Sparse MoE} & \textbf{Sparse MoE} \\
                                  & \textit{t=1} & \textit{t=0.7}     & \textit{t=0.3}     & \textit{}       & \textit{t=1} & \textit{t=0.7}      & \textit{t=0.3}      \\ \hline
C1862$_\text{Ki}$                    & \textbf{0.772$\pm$0.01}  & 0.77$\pm$0.01         & 0.760$\pm$0.01        & 0.771$\pm$0.0      & 0.767$\pm$0.01  & 0.751$\pm$0.02         & 0.708$\pm$0.03         \\
C1871$_\text{Ki}$                   & 0.505$\pm$0.01  & 0.54$\pm$0.01         & \textbf{0.549$\pm$0.01}        & 0.539$\pm$0.01     & 0.532$\pm$0.03  & 0.509$\pm$0.03         & 0.454$\pm$0.04         \\
C2034$_\text{Ki}$                    & 0.492$\pm$0.01  & 0.52$\pm$0.02         & 0.523$\pm$0.04        & 0.508$\pm$0.02     & 0.530$\pm$0.03  & \textbf{0.533$\pm$0.02}         & 0.475$\pm$0.02         \\
C2047$_\text{EC}$                  & 0.245$\pm$0.05  & \textbf{0.32$\pm$0.05}         & 0.204$\pm$0.01        & 0.209$\pm$0.03     & 0.239$\pm$0.08  & 0.253$\pm$0.02         & 0.232$\pm$0.04         \\
C204$_\text{Ki}$                     & 0.796$\pm$0.01  & 0.79$\pm$0.01         & \textbf{0.798$\pm$0.01}        & 0.786$\pm$0.0      & 0.786$\pm$0.01  & 0.757$\pm$0.01         & 0.776$\pm$0.02         \\
C2147$_\text{Ki}$                    & \textbf{0.876$\pm$0.01}  & 0.87$\pm$0.0          & 0.873$\pm$0.01        & 0.856$\pm$0.01     & 0.861$\pm$0.01  & 0.865$\pm$0.01         & 0.851$\pm$0.01         \\
C214$_\text{Ki}$                     & \textbf{0.654$\pm$0.01}  & 0.64$\pm$0.01         & 0.615$\pm$0.01        & 0.630$\pm$0.03     & 0.634$\pm$0.03  & 0.643$\pm$0.02         & 0.627$\pm$0.01         \\
C218$_\text{Ki}$                   & 0.582$\pm$0.02  & 0.58$\pm$0.04         & \textbf{0.585$\pm$0.02}        & 0.581$\pm$0.01     & 0.573$\pm$0.0   & 0.570$\pm$0.02         & 0.583$\pm$0.01         \\
C219$_\text{Ki}$                     & \textbf{0.398$\pm$0.01}  & 0.31$\pm$0.04         & 0.383$\pm$0.02        & 0.349$\pm$0.01     & 0.358$\pm$0.02  & 0.331$\pm$0.03         & 0.332$\pm$0.01         \\
C228$_\text{Ki}$                     & \textbf{0.694$\pm$0.01}  & 0.68$\pm$0.01         & 0.648$\pm$0.0         & 0.692$\pm$0.02     & 0.657$\pm$0.02  & 0.662$\pm$0.05         & 0.661$\pm$0.03         \\
C231$_\text{Ki}$                     & 0.729$\pm$0.01  & 0.74$\pm$0.0          & 0.734$\pm$0.01        & \textbf{0.751$\pm$0.01}     & 0.733$\pm$0.02  & 0.723$\pm$0.02         & 0.715$\pm$0.02         \\
C233$_\text{Ki}$                     & 0.708$\pm$0.01  & \textbf{0.71$\pm$0.01}         & 0.672$\pm$0.02        & 0.707$\pm$0.01     & 0.701$\pm$0.02  & 0.687$\pm$0.02         & 0.652$\pm$0.01         \\
C234$_\text{Ki}$                     & 0.669$\pm$0.01  & 0.65$\pm$0.0          & 0.668$\pm$0.01        & 0.682$\pm$0.02     & 0.656$\pm$0.02  & \textbf{0.684$\pm$0.02}         & 0.650$\pm$0.02         \\
C235$_\text{EC}$                   & 0.665$\pm$0.01  & \textbf{0.7$\pm$0.02}          & 0.651$\pm$0.01        & 0.690$\pm$0.01     & 0.675$\pm$0.01  & 0.694$\pm$0.0          & 0.699$\pm$0.02         \\
C236$_\text{Ki}$                     & \textbf{0.750$\pm$0.0}   & 0.74$\pm$0.01         & 0.741$\pm$0.01        & 0.731$\pm$0.0      & 0.707$\pm$0.01  & 0.711$\pm$0.01         & 0.707$\pm$0.01         \\
C237$_\text{EC}$                   & \textbf{0.671$\pm$0.03}  & 0.66$\pm$0.01         & 0.658$\pm$0.0         & 0.669$\pm$0.01     & 0.651$\pm$0.02  & 0.627$\pm$0.01         & 0.648$\pm$0.03         \\
C237$_\text{Ki}$                     & \textbf{0.724$\pm$0.0}   & 0.71$\pm$0.0          & 0.716$\pm$0.0         & 0.706$\pm$0.01     & 0.695$\pm$0.01  & 0.707$\pm$0.02         & 0.708$\pm$0.01         \\
C238$_\text{Ki}$                     & 0.681$\pm$0.01  & 0.68$\pm$0.2          & 0.692$\pm$0.01        & \textbf{0.703$\pm$0.01}     & 0.682$\pm$0.01  & 0.658$\pm$0.02         & 0.625$\pm$0.03         \\
C239$_\text{EC}$                   & 0.445$\pm$0.02  & \textbf{0.58$\pm$0.02}         & 0.436$\pm$0.01        & 0.465$\pm$0.01     & 0.472$\pm$0.04  & 0.496$\pm$0.02         & 0.511$\pm$0.03         \\
C244$_\text{Ki}$                     & 0.809$\pm$0.01  & \textbf{0.81$\pm$0.0}          & 0.807$\pm$0.0         & 0.802$\pm$0.0      & 0.794$\pm$0.01  & 0.786$\pm$0.01         & 0.781$\pm$0.01         \\
C262$_\text{Ki}$                     & 0.594$\pm$0.01  & 0.6$\pm$0.03          & \textbf{0.610$\pm$0.02}        & \textbf{0.610$\pm$0.01}     & 0.515$\pm$0.07  & 0.584$\pm$0.01         & 0.577$\pm$0.07         \\
C264$_\text{Ki}$                     & 0.681$\pm$0.01  & \textbf{0.7$\pm$0.0}           & 0.688$\pm$0.01        & 0.661$\pm$0.01     & 0.692$\pm$0.01  & 0.685$\pm$0.01         & 0.645$\pm$0.02         \\
C2835$_\text{Ki}$                    & 0.795$\pm$0.01  & 0.76$\pm$0.01         & 0.762$\pm$0.01        & 0.789$\pm$0.0      & 0.777$\pm$0.01  & \textbf{0.797$\pm$0.0}          & 0.791$\pm$0.0          \\
C287$_\text{Ki}$                     & 0.509$\pm$0.01  & 0.5$\pm$0.01          & 0.493$\pm$0.0         & 0.471$\pm$0.02     & \textbf{0.523$\pm$0.02}  & 0.508$\pm$0.02         & 0.455$\pm$0.02         \\
C2971$_\text{Ki}$                    & 0.848$\pm$0.02  & 0.85$\pm$0.02         & 0.820$\pm$0.01        & \textbf{0.861$\pm$0.0}      & 0.820$\pm$0.02  & 0.792$\pm$0.02         & 0.792$\pm$0.02         \\
C3979$_\text{EC}$                  & 0.500$\pm$0.06  & 0.48$\pm$0.04         & \textbf{0.544$\pm$0.03}        & 0.537$\pm$0.02     & 0.526$\pm$0.05  & 0.539$\pm$0.02         & 0.490$\pm$0.03         \\
C4005$_\text{Ki}$                    & 0.583$\pm$0.01  & 0.55$\pm$0.02         & \textbf{0.608$\pm$0.01}        & 0.601$\pm$0.03     & 0.548$\pm$0.03  & 0.541$\pm$0.0          & 0.558$\pm$0.02         \\
C4203$_\text{Ki}$                    & \textbf{0.384$\pm$0.02}  & 0.38$\pm$0.04         & 0.335$\pm$0.04        & 0.293$\pm$0.04     & 0.292$\pm$0.06  & 0.300$\pm$0.01         & 0.284$\pm$0.05         \\
C4616$_\text{EC}$                  & 0.458$\pm$0.04  & 0.49$\pm$0.02         & \textbf{0.529$\pm$0.06}        & 0.467$\pm$0.02     & 0.503$\pm$0.03  & 0.470$\pm$0.03         & 0.452$\pm$0.04         \\
C4792$_\text{Ki}$                    & 0.533$\pm$0.01  & 0.52$\pm$0.0          & 0.544$\pm$0.02        & \textbf{0.548$\pm$0.01}     & 0.486$\pm$0.04  & 0.537$\pm$0.01         & 0.521$\pm$0.03         \\ \hline
Average                           & 0.6250       & \textbf{0.6279}             & 0.6215             & 0.6222          & 0.6128       & 0.6133              & 0.5987              \\                                                       \Xhline{1.5pt} 
\end{tabular}
\end{adjustbox}
\end{table}
\end{landscape}
